\documentclass[10pt,journal,compsoc]{IEEEtran}

\usepackage{arydshln}
\usepackage{amsmath,amsfonts,amssymb}
\usepackage{algorithmic}
\usepackage{algorithm}
\usepackage{array}
\usepackage{subfigure}
\usepackage{graphicx}
\usepackage{textcomp}
\usepackage{stfloats}
\usepackage{url}
\usepackage{verbatim}
\usepackage{graphicx}
\usepackage{color}
\usepackage{multirow}
\usepackage{booktabs}

\usepackage{bm}

\usepackage{xcolor}
\usepackage{hyperref}
\usepackage{makecell}

\makeatletter
\newcommand*{\rom}[1]{\expandafter\@slowromancap\romannumeral #1@}
\makeatother

\usepackage{colortbl}

\definecolor{tablegray}{gray}{.9}
\newcommand{\etal}{\textit{et al}.}
\newcommand{\ie}{\textit{i}.\textit{e}.}
\newcommand{\eg}{\textit{e}.\textit{g}.}

\ifCLASSOPTIONcompsoc

  \usepackage[nocompress]{cite}
\else
  \usepackage{cite}
\fi

\ifCLASSINFOpdf
 
\else
 
\fi

\begin{document}

\title{Analysis of Video Quality Datasets  via Design of Minimalistic Video Quality Models}

\author{Wei~Sun,
        Wen~Wen,
        Xiongkuo~Min,~\IEEEmembership{Member,~IEEE,}
        Long~Lan,
        Guangtao~Zhai,~\IEEEmembership{Senior Member,~IEEE,}
        and Kede~Ma,~\IEEEmembership{Senior Member,~IEEE}

\IEEEcompsocitemizethanks{\IEEEcompsocthanksitem Wei Sun and Xiongkuo Min are with the Institute of Image Communication and Information Processing, Shanghai Jiao Tong University, Shanghai 200240, China (e-mail: \{sunguwei, minxiongkuo\}@sjtu.edu.cn).
\IEEEcompsocthanksitem Wen Wen is with the Department of Computer Science, City University of Hong Kong, Kowloon, Hong Kong (e-mail: wwen29-c@my.cityu.edu.hk).
\IEEEcompsocthanksitem Long Lan is with the Institute for Quantum Information \& State Key Laboratory of High Performance Computing, College of Computer Science and Technology, National University of Defense Technology, Changsha 410073, China (e-mail: long.lan@nudt.edu.cn).
\IEEEcompsocthanksitem Guangtao Zhai is with the Institute of Image Communication and Information Processing, and also with the MoE Key Lab of Artificial Intelligence, AI Institute, Shanghai Jiao Tong University, Shanghai 200240, China (e-mail: zhaiguangtao@sjtu.edu.cn).
\IEEEcompsocthanksitem Kede Ma is with the Department of Computer Science, and also with the Shenzhen Research Institute, City University of Hong Kong, Kowloon, Hong Kong (e-mail: kede.ma@cityu.edu.hk).
}
\thanks{This work was supported in part by the National Natural Science Foundation of China under Grants 62071407, 62225112, 62301316, 62376282 and 62271312,  the Fundamental Research Funds for the Central Universities, the National Key R\&D Program of China (2021YFE0206700), the Science and Technology Commission of Shanghai Municipality (2021SHZDZX0102), the Shanghai Committee of Science and Technology (22DZ2229005), the China Postdoctoral Science Foundation under Grants 2023TQ0212 and 2023M742298, the Postdoctoral Fellowship Program of CPSF under Grant GZC20231618, and the CCF-Tencent Rhino Bird Fund (9239041).}
\thanks{Corresponding Authors: Xiongkuo Min and Guangtao Zhai.}
}

\IEEEtitleabstractindextext{%
\begin{abstract}
Blind video quality assessment (BVQA) plays an indispensable role in monitoring and improving the end-users' viewing experience in various real-world video-enabled media applications. As an experimental field, the improvements of BVQA models have been measured primarily on a few human-rated VQA datasets. Thus, it is crucial to gain a better understanding of existing VQA datasets in order to properly evaluate the current progress in BVQA. Towards this goal, we conduct a first-of-its-kind computational analysis of VQA datasets via designing minimalistic BVQA models. By minimalistic, we restrict our family of BVQA models to build only upon basic blocks: a video preprocessor (for aggressive spatiotemporal downsampling), a spatial quality analyzer, an optional temporal quality analyzer, and a quality regressor, all with the simplest possible instantiations. By comparing the quality prediction performance of different model variants on eight VQA datasets with realistic distortions, we find that nearly all datasets suffer from the easy dataset problem of varying severity, some of which even admit blind image quality assessment (BIQA) solutions.
We additionally justify our claims by comparing our model generalization capabilities on these VQA datasets, and by ablating a dizzying set of BVQA design choices related to the basic building blocks. Our results cast doubt on the current progress in BVQA, and meanwhile shed light on good practices of constructing next-generation VQA datasets and models. Code is available at \url{https://github.com/sunwei925/MinimalisticVQA.git}.
\end{abstract}
}

\maketitle

\IEEEdisplaynontitleabstractindextext

\IEEEpeerreviewmaketitle

\ifCLASSOPTIONcompsoc
\IEEEraisesectionheading{\section{Introduction}\label{sec:introduction}}
\else
\section{Introduction}
\label{sec:introduction}
\fi

\IEEEPARstart{W}{e} are undoubtedly living in an era, where we are exposed to a great volume of video data through various video on demand and live streaming media applications. No matter at what stages along the video delivery chain from video creation to consumption, the perceptual video quality is of central concern \cite{juluri2015measurement}.

Video quality assessment (VQA) plays a fundamental role in a wide range of video processing systems, such as filtering videos of extremely low visual quality from the content producer side, guiding video compression and transmission algorithms to achieve perceptually optimal rate-distortion trade-off, and ensuring smooth and faithful rendering on various displays of different spatial resolutions, dynamic ranges, and color gamuts. How to perform reliable VQA subjectively~\cite{seshadrinathan2010study, nuutinen2016cvd2014, ghadiyaram2017capture, hosu2017konstanz, sinno2018large, wang2019youtube, chen2019qoe, ying2021patch, yu2022subjective} and objectively~\cite{saad2014blind, mittal2015completely, liu2018end, korhonen2019two, li2019quality, korhonen2020blind, tu2021ugc, ying2021patch, yu2021predicting, yi2021attention, wang2021rich, li2022blindly, sun2022deep, wu2022fast, wu2022disentangling} present two challenging problems that have been extensively studied in the interdisciplinary fields of signal and image processing, psychophysics, computational neuroscience, and machine learning.

Subjective VQA involves two key steps: \textit{sample selection} and \textit{subjective testing}, which outputs a video dataset with perceptual quality annotations in the form of \textbf{m}ean \textbf{o}pinion \textbf{s}core\textbf{s} (MOSs). During sample selection, the visual distortion types and levels are largely determined by the video applications of interest. Early subjective VQA~\cite{seshadrinathan2010study, li2019avc, mackin2015study, lee2021subjective} focuses on synthetic visual distortions arising from different video processing stages, including spatiotemporal downsampling, compression, and transmission. Recent subjective VQA~\cite{nuutinen2016cvd2014, ghadiyaram2017capture, hosu2017konstanz, sinno2018large, wang2019youtube, chen2019qoe, ying2021patch, yu2022subjective} shifts the attention to realistic visual distortions arising during video capture (and subsequent realistic video processing over the Internet). To encourage wide content coverage, Vonikakis~\etal~\cite{vonikakis2016shaping} cast sample selection as a mixed integer linear program to enforce specific marginal distributions over different video attributes (\eg, spatial and temporal information). This dataset shaping technique has inspired later researchers to give a careful treatment of sample selection during dataset construction~\cite{hosu2017konstanz,wang2019youtube,ying2021patch}.
With respect to subjective testing methodologies, International Telecommunication Union (ITU) has made several recommendations~\cite{recommendation910subjective,union2016methods,series2012methodology} regarding the experimental environment, the video clip length, the number of subjects, and the rating type.

Objective VQA aims to build computational models to accurately predict the human perception of video quality. When the reference video is assumed, full-reference VQA models~\cite{wang2004image,moorthy2010efficient,zhang2018unreasonable,ding2020image,li2018vmaf} can be constructed to compute the perceptual quality of a test video by comparing it to the corresponding reference. A more practical yet challenging scenario is to design VQA models~\cite{saad2014blind, korhonen2019two, li2019quality, wu2022fast} with only reliance on the test video itself, hence the name no-reference VQA or blind VQA (BVQA). In most Internet-based video applications, the reference video is not practically accessible, which makes BVQA the focus of this paper. Compared to blind image quality assessment (BIQA), BVQA is more difficult due to the added temporal dimension, which complicates the computational modeling of visual distortions possibly localized in space and time.

As an experimental field, the improvements of BVQA models have been evaluated routinely on a few human-rated VQA datasets. To probe how reliable current progress in BVQA is, we need to gain a better understanding of these VQA datasets. 
Towards this goal, we conduct one of the first computational analyses of existing VQA datasets~\cite{nuutinen2016cvd2014,ghadiyaram2017capture,hosu2017konstanz,sinno2018large,wang2019youtube,chen2019qoe,ying2021patch,yu2022subjective}, with emphasis on identifying the \textit{easy dataset} problem~\cite{cao2024image}. We say that a VQA dataset suffers from the easy dataset problem if it
\begin{itemize}
    \item admits close-to-BIQA solutions,
    \item or poses little challenge to current BVQA models.
\end{itemize}
The first criterion indicates that temporal distortions (if any) in
 the VQA dataset are dominated by or strongly correlated with spatial distortions. As a result, a BIQA model that properly penalizes spatial distortions tends to perform well. This goes against the original intent of constructing the VQA dataset.
 The second criterion suggests that the VQA dataset exposes few failure cases of current BVQA models, leading to a significant waste of the human labeling budget that could have been reallocated to more difficult videos.

\begin{table*}
	\centering
	\renewcommand{\arraystretch}{1.25}
	\caption{Summary of eight VQA datasets.  Duration is in seconds}
	\label{overview_vqa_dataset}
 
	\begin{tabular}{lcccccccc}
	\toprule[.15em]
		\multirow{1}{*}{Dataset} & \multirow{1}{*}{Year} &\multirow{1}{*}{\# of Videos} & \multirow{1}{*}{\# of Scenes} & \multirow{1}{*}{Resolution} & Duration& \multirow{1}{*}{Frame Rate} &  \multirow{1}{*}{Type} & \multirow{1}{*}{Environment} \\
		\hline
  	CVD2014	\cite{nuutinen2016cvd2014} &2014& 234 & 5& 480p, 720p & 10-25& 10-32 &  In-captured &  In-laboratory \\
        LIVE-Qualcomm \cite{ghadiyaram2017capture} &2017& 208 &54 & 1080p & 15&  30&  In-captured & In-laboratory \\
		KoNViD-1k \cite{hosu2017konstanz} &2017& 1,200 & 1,200 & 540p & 8 & 24, 25, 30 &  Internet-collected & Crowdsourcing \\
  	LIVE-VQC \cite{sinno2018large} &2018& 585 & 585 & 240p-1080p & 10& 30 &  In-captured & Crowdsourcing \\
		YouTube-UGC \cite{wang2019youtube} &2019& 1,500 & 1,500 & 360p-4K & 20 & 30 &  Internet-collected & Crowdsourcing \\
		LBVD \cite{chen2019qoe} &2019& 1,013 & 1,013 & 240p-540p & 10 & $<$ 30 &  Internet-collected &  In-laboratory \\
		LSVQ \cite{ying2021patch} &2021& 38,811 & 38,811 & 99p-4K & 5-12 & $<$ 60 &  Internet-collected & Crowdsourcing \\
		LIVE-YT-Gaming \cite{yu2022subjective} &2022& 600& 600& 360p-1080p& 8-9& 30, 60&  Internet-collected & Crowdsourcing \\
    \bottomrule[.15em]
	\end{tabular}
\end{table*}

Our main computational analysis of VQA datasets examines the first criterion of the easy dataset problem. Specifically, we design a family of minimalistic BVQA models, which consist of four basic building blocks: a video preprocessor, a spatial quality analyzer, an optional temporal quality analyzer, and a quality regressor. The video preprocessor is designed to aggressively downsample the input test video in both spatial and temporal dimensions, giving rise to a few key video frames. The spatial quality analyzer acts as a BIQA model to compute a spatial quality representation for each key frame. The temporal quality analyzer is optional, extracting a temporal quality representation from a video chunk centered at each key frame. The quality regressor is responsible for computing and pooling local quality scores from the concatenated spatial and temporal features to obtain a global quality score for the test video. The four building blocks can be instantiated with different computational structures, equipped with different quality-aware initializations, and end-to-end optimized.

 We train ten BVQA model variants on eight VQA datasets~\cite{nuutinen2016cvd2014,ghadiyaram2017capture,hosu2017konstanz,sinno2018large,sinno2018large,wang2019youtube,chen2019qoe,ying2021patch,yu2022subjective} by minimizing the Pearson linear correlation coefficient (PLCC) between  MOSs and model predictions. Provided sufficient optimization, we attribute the performance variations to the implementation differences of the basic building blocks, especially the spatial and temporal quality analyzers. The primary observation from our computational analysis is that nearly all datasets suffer from the easy dataset problem by satisfying the first criterion to varying degrees. 
To further support our claims, we examine the second criterion of the easy dataset problem by testing the generalization of our models trained on the largest VQA dataset~\cite{ying2021patch} to the other seven datasets. We additionally ablate 
a dizzying set of BVQA design choices related to the basic building blocks.

In summary, our contributions include
\begin{itemize}
\item a family of minimalistic BVQA models, serving as strong baselines and laying the foundation for analyzing VQA datasets,
\item a computational analysis of eight VQA datasets, which we are able to empirically rank according to the severity of the easy dataset problem,
\item a proposal to construct more useful VQA datasets from the sample selection and subjective testing perspectives. 
\end{itemize}

\section{Related Work}
\label{review}
In this section, we provide an overview of eight VQA datasets with realistic visual distortions. We then review and characterize existing BVQA methods, putting our minimalistic BVQA models in proper context.

\begin{figure*}[t]
\centering
\subfigure[ CVD2014]
{\label{fig:CVD2014}
\includegraphics[width=0.225\textwidth]{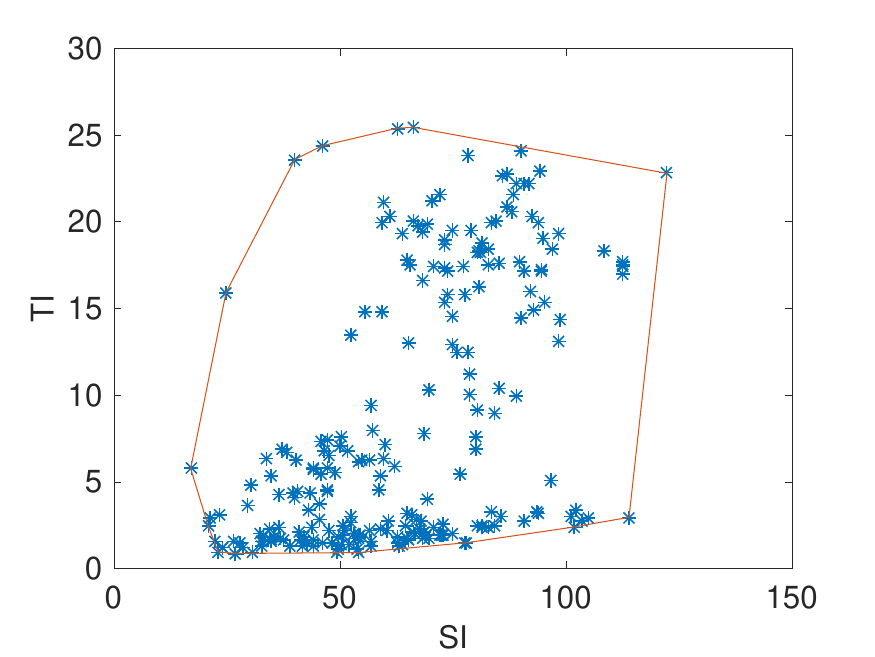}
}
\subfigure[ LIVE-Qualcomm]
{\label{fig:LIVE_Qualcomm}
\includegraphics[width=0.225\textwidth]{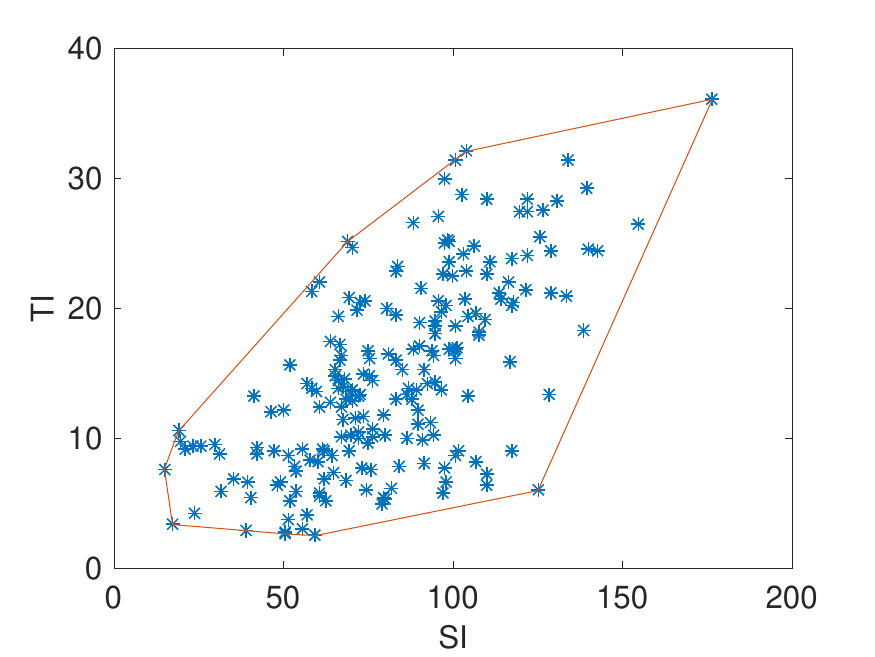}
}
\subfigure[ KoNViD-1k]
{\label{fig:KoNViD_1k}
\includegraphics[width=0.225\textwidth]{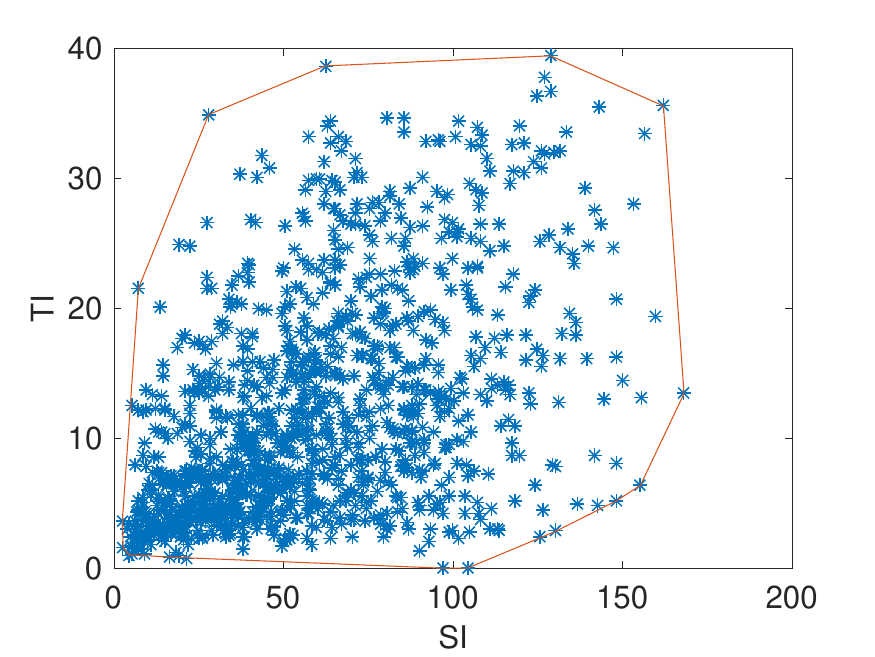}
}
\subfigure[ LIVE-VQC]
{\label{fig:LIVE_VQC}
\includegraphics[width=0.225\textwidth]{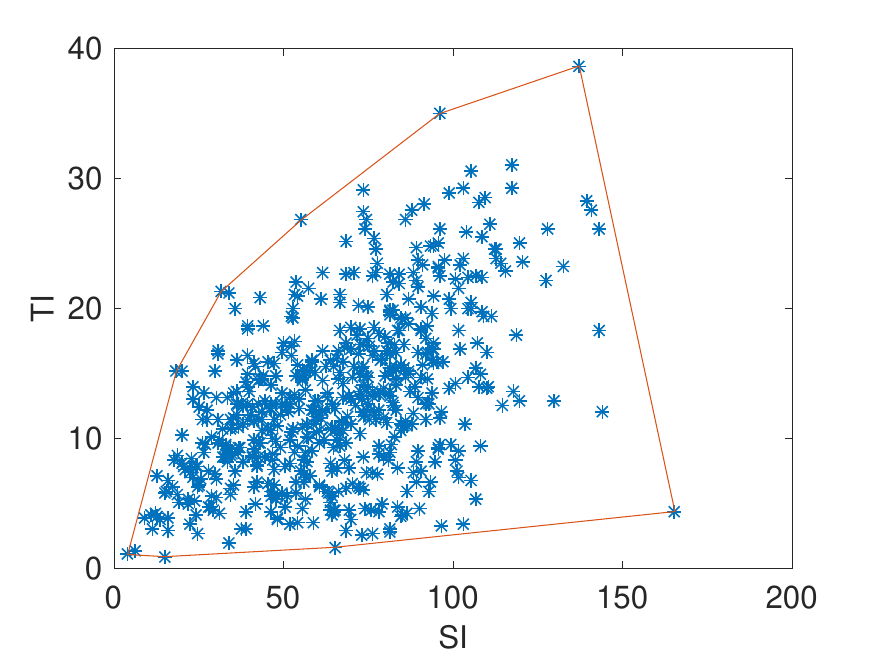}
}
\subfigure[ YouTube UGC]
{\label{fig:youtube_ugc}
\includegraphics[width=0.225\textwidth]{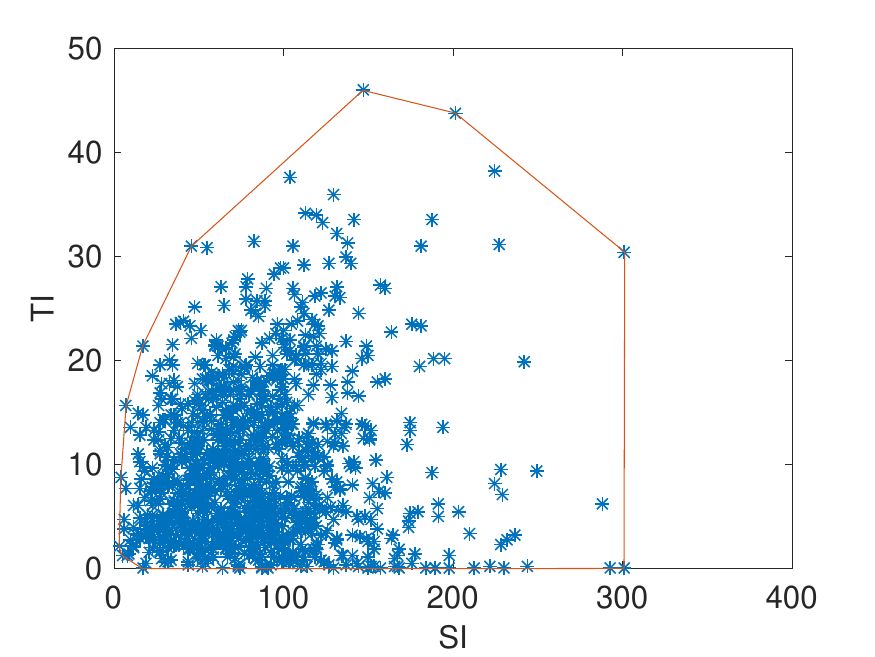}
}
\subfigure[ LBVD]
{\label{fig:LBVD}
\includegraphics[width=0.225\textwidth]{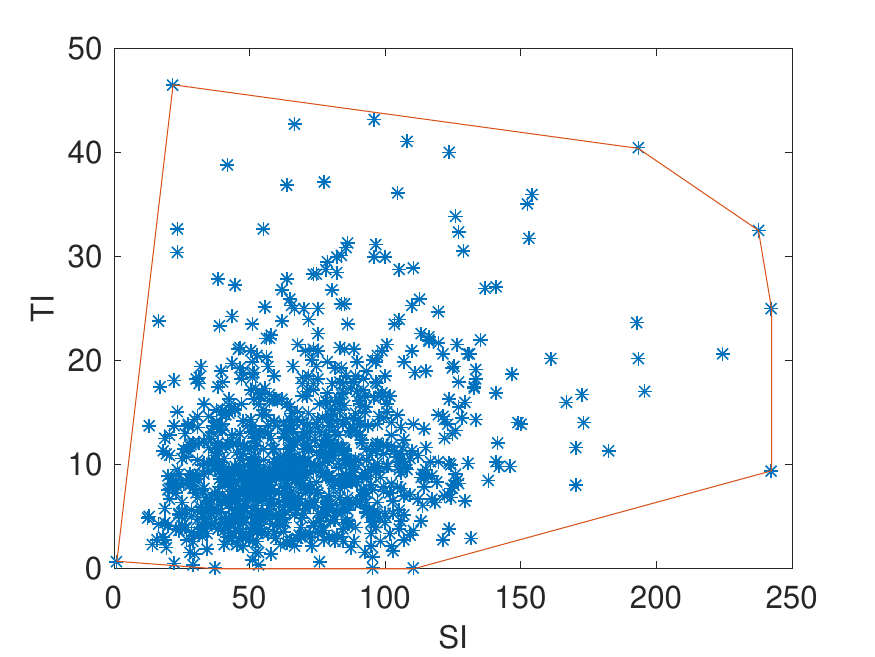}
}
\subfigure[ LSVQ]
{\label{fig:LSVQ}
\includegraphics[width=0.225\textwidth]{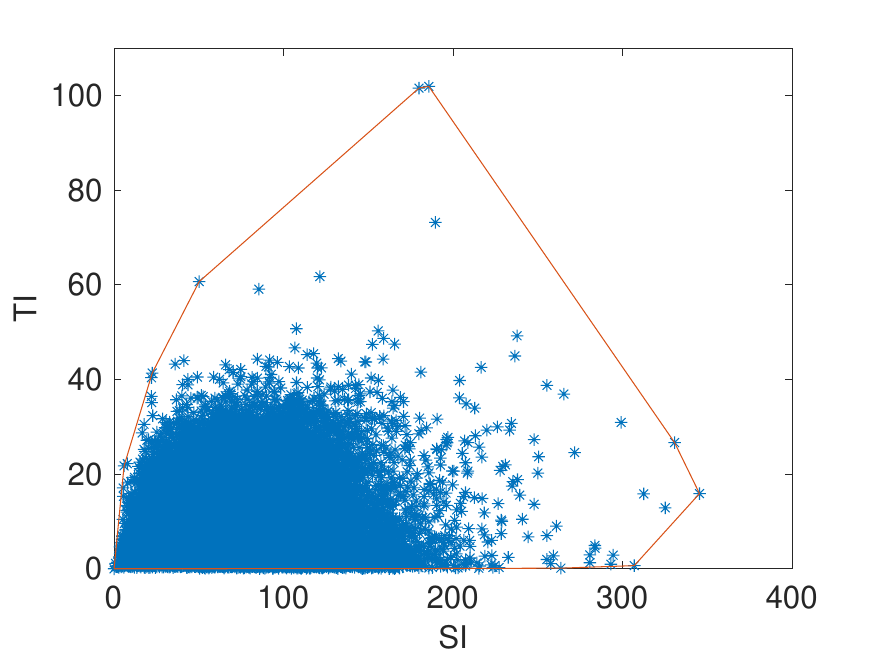}
}
\subfigure[ LIVE-YT-Gaming]
{\label{fig:LIVEYTGaming}
\includegraphics[width=0.225\textwidth]{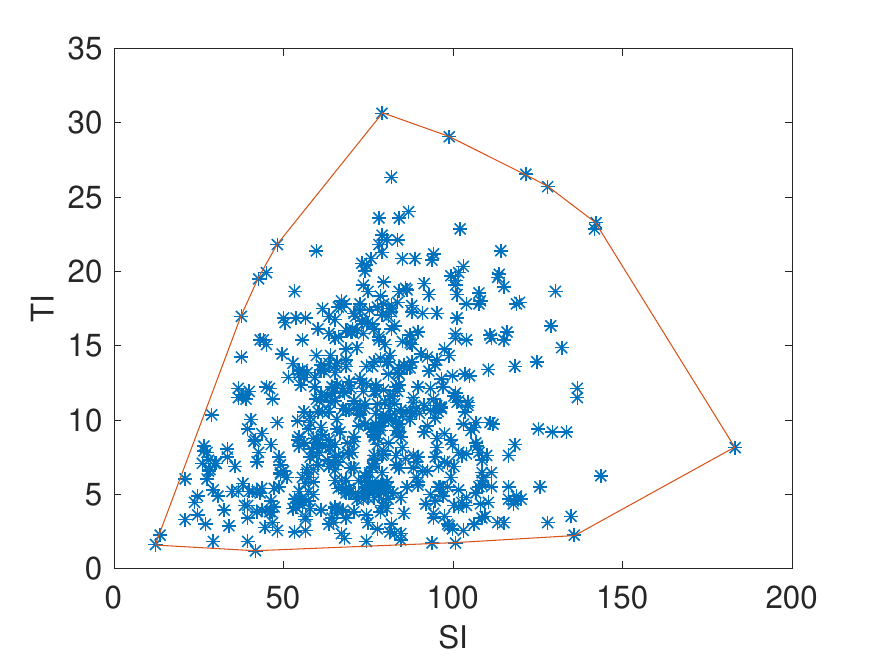}
}
\caption{Scatter plots of spatial information (SI) versus temporal information (TI) of the eight VQA datasets with the corresponding convex hulls. Each dot represents a video.}
\label{fig:SI_TI_figure}
\end{figure*}

\subsection{VQA Datasets}
\label{vqa_dataset}
It is common to divide the videos with realistic distortions into two categories: \textit{in-captured videos}, which are directly captured by various cameras with minimal subsequent processing (\eg, transcoding and transmission) and \textit{Internet-collected videos}, which suffer from a greater mixture of video distortions along the video delivery chain.
\subsubsection{In-Captured VQA Datasets}
\textbf{CVD2014} \cite{nuutinen2016cvd2014} presents the first subjective quality study of in-captured videos, containing $234$ videos recorded by $78$ cameras with varying compression formats, spatial resolutions, and frame rates. In addition to MOSs, detailed video attributes such as sharpness, saturation, color pleasantness, lightness, motion fluency, and sound quality were also rated. Nevertheless, CVD2014 only includes five hand-picked scenes captured by nearly still cameras, making it little representative of 
hard natural videos encountered in the real world.

To increase scene diversity, \textbf{LIVE-Qualcomm} \cite{ghadiyaram2017capture} was created to comprise $208$ videos using $8$ different smartphones under $54$ unique scenes\footnote{Compared to CVD2014, LIVE-Qualcomm decreases the number of captured devices while increases the number of scenes.}. The dataset is further divided into six subsets, each corresponding to one dominant distortion type, characterized by spatial artifacts (\eg, noise and blockiness), color, exposure, autofocus, sharpness, and stabilization.  
The subjective experiments of CVD2014 and LIVE-Qualcomm were both conducted under well-controlled laboratory environments, which limits the total number of subjects to be recruited and videos to be rated.

\textbf{LIVE-VQC} \cite{sinno2018large} consists of $585$ in-captured videos by $80$ mobile cameras, each  corresponding to a unique scene. LIVE-VQC spans a wide range of natural scenes under different lighting conditions as well as diverse levels of motion. There are $18$ video resolutions, ranging from 240p to 1080p.  The subjective user study was conducted on the Amazon Mechanical Turk, in which $4,776$ subjects took part. Compared with a standard laboratory environment where the display has a fixed and high resolution, the viewing conditions (such as the effective viewing distance and the ambient luminance) of crowdsourcing platforms vary greatly. For example, the test video, regardless of its actual resolution, is adapted to the resolution of the screen in the crowdsourcing experiment (which is typically larger than $1,280\times720$). As a consequence, the collected MOSs may not account for the influence of spatial resolutions.

\subsubsection{Internet-Collected VQA Datasets}
\label{internet_collected_video}

\textbf{KoNViD-1k} \cite{hosu2017konstanz} is the first dataset to study the visual quality of Internet videos. A density-based fair sampling strategy was used to select $1,200$ videos with the resolution of $960\times 540$ from YFCC100M~\cite{thomee2016yfcc100m} based on six attributes: blur, colorfulness, contrast, spatial information, temporal information, and predicted video quality by V-NIQE~\cite{mittal2012making}. 
The MOSs were also crowdsourced.  Originated from YFCC100M, a Flickr dataset containing videos uploaded from 2004 to 2014, most videos in KoNViD-1k are dominated by spatial distortions like noise, blur, low visibility, and compression artifacts.

To enrich video content, \textbf{YouTube-UGC} \cite{wang2019youtube} selects an initial set of $1.5$ million videos belonging to $15$ overlapping categories\footnote{These are Animation, Cover Song, Gaming, HDR, How-To, Lecture, Live Music, Lyric Video, Music Video, News Clip, Sports, Television Clip, Vertical Video, Vlog, and VR.} from YouTube with resolutions spanning from  360p to 4K. Similar to KoNViD-1k, YouTube-UGC relies on a (quantized) density-based sampling strategy to select $1,500$ videos\footnote{Up to now, only 1,228 videos are publicly available due to copyright issues.} with respect to spatial, color, temporal, and chunk variation attributes (measured by compression bitrates). Compared with other datasets, a portion of  YouTube-UGC videos are of higher quality. Meanwhile, the time duration of each video is of 20-second long with possible scene changes.

\textbf{LBVD} \cite{chen2019qoe} gathers  $1,013$ live broadcasting videos based on HTTP adaptive streaming. The main sources of distortions include unstable network conditions (\eg, rebuffering and stalling), camera shake (\eg, jittering and motion blur), and drastic dynamic range changes. The videos are mostly human-centric under different lighting conditions. The predominant resolutions are 240p, 360p, and 540p, which are significantly lower compared to other datasets. The subjective experiment was conducted in a well-controlled laboratory environment.

Till then, the number of included videos is rather limited, which is in the order of hundreds. To overcome this shortcoming, \textbf{LSVQ} \cite{ying2021patch} was constructed with $38,811$ videos sampled from the Internet Archive and YFCC100M by matching six video feature distributions\footnote{These are luminance, colorfulness, luminance contrast, number of detected faces, spatial Gaussian derivative filter responses, and temporal Gaussian derivatives.} \cite{vonikakis2016shaping}. LSVQ covers a broad range of resolutions and frame rates. Because of the large number of videos (including three variants of video patches) that need to be rated, the subjective experiment was practically held online. LSVQ enables training BVQA models for local video quality prediction, but is beyond the scope of the current paper.

\textbf{LIVE-YT-Gaming} \cite{yu2022subjective} is a VQA dataset targeting streamed gaming videos.  A total of $600$ gaming videos belonging to $59$ games were obtained either from the Internet Archive or by direct recording using professional software. Since gaming videos are synthetically rendered, the distortions pertain to 
the graphic configuration settings of the operating system, the settings built in each game, and the grade of the recording software. Hence, both the content statistics, as well as the distortion appearances, are significantly different from those of natural videos captured by optical cameras. For example, gaming videos are generally composed of sharper edges with minimal noise, but may exhibit loss of details in grass and water regions, unnaturalness of shadows, and aliasing. 

Table~\ref{overview_vqa_dataset} gives a summary of the eight VQA datasets. From the \textit{sample selection} perspective, many prefer to incorporate diverse videos with respect to several video attributes. Nevertheless, diversity is just one piece of the story; the included videos shall also be \textit{difficult} to challenge BVQA models, which is unfortunately overlooked, causing the easy dataset problem. From the \textit{subjective testing} perspective, crowdsourcing has been proven to collect reliable MOSs provided that the psychophysical experimental procedure is designed appropriately. And thus we assume the reliability of MOSs in all datasets without discussing the label noise and uncertainty in this paper.

Previous studies~\cite{winkler2012analysis,robitza2021impact} analyzed VQA datasets through a set of low-level video descriptors, in particular the spatial information (SI) and the temporal information (TI)~\cite{recommendation910subjective}. 
Fig.~\ref{fig:SI_TI_figure} shows such scatter plots of SI versus TI of the eight VQA datasets, from which we may peek into some spatiotemporal characteristics of these datasets. For instance, LIVE-VQC and LIVE-Qualcomm have the highest average TI values, while the latter possesses the highest linear relationship between SI and TI.   However, it is important to highlight the limitations of SI and TI. Specifically, the Sobel operator~\cite{parker2010algorithms} used to compute SI is noise-sensitive. The frame difference in TI is sensitive to tiny geometric distortions (such as translation, rotation, and dilation) that are otherwise imperceptible to the human eye. Therefore, relying solely on SI and TI is limited in characterizing the spatiotemporal statistics of VQA datasets, let alone relating such statistics to perceived visual quality. Our computational analysis provides a more comprehensive and direct understanding of VQA dataset characteristics from a quality assessment perspective.

\begin{table*}
	\centering
	\renewcommand{\arraystretch}{1.25}
	\caption{Categorization of DNN-based BVQA models based on the four basic building blocks and optimization strategies. ``None'' in Video Preprocessor column means that no such preprocessing, while ``---'' indicates no such building block is incorporated into the corresponding BVQA model. PaQ-2-PiQ~\cite{ying2020patches} and UNIQUE~\cite{zhang2021uncertainty} are two BIQA models that pre-trained ResNet-18~\cite{he2016deep} and ResNet-50~\cite{he2016deep} on the LIVE-FB~\cite{ying2020patches} dataset and four merged IQA datasets, respectively. EfficientNet-B0~\cite{tan2019efficientnet} in CoINVQ was pre-trained on tasks involving video classification and distortion type classification, whereas D3D~\cite{stroud2020d3d} in CoINVQ was pre-trained for video compression classification. ConvNeXt-T~\cite{liu2022convnet} in DOVER was initially pre-trained for image aesthetics assessment with the cross-scale feature consistency constraint (not shown in the Loss Function column due to the space limit). 
 FAST-VQA and DOVER sample a fixed number of frames (\ie, 32) regardless of the video duration in the VQA datasets, and we convert it to frames per second (fps)}
	\label{tab:bvqa_models}

    \resizebox{1\textwidth}{!}{
	\begin{tabular}{lccccccc}
	\toprule[.15em]
		\multirow{2}{*}{Model} & \multicolumn{2}{c}{Video Preprocessor} & Spatial Quality & Temporal Quality & Quality & \multirow{2}{*}{Optimization Strategy}& \multirow{2}{*}{Loss Function} \\
		 & Spatial Branch & Temporal Branch & Analyzer (SA)  & Analyzer (TA) &  Regressor (QR) & &  \\
		\hline
  	VSFA \cite{li2019quality} &  None & --- & ResNet-50 & --- & GRU & Fix SA, Train QR &  L1 \\
    Yi21 \cite{yi2021attention} &  $448 \times 448$, $1$ fps& ---& VGG16 & --- & GRU & End-to-End Training &  L1 \\
    PatchVQ \cite{ying2021patch} &  None &None& PaQ-2-PiQ &  3D ResNet-18 & InceptionTime & Fix SA\&TA, Train QR &  L1 \\
    CoINVQ \cite{wang2021rich} & $360\times 640$, $1$ fps & $180\times320$, $5$ fps& EfficientNet-B0 &  D3D & MLP & Fix SA\&TA, Train QR &  L1 \\
    Li22 \cite{li2022blindly} &  None &None& UNIQUE &  SlowFast & GRU & Fix SA\&TA, Train QR &  PLCC\&SRCC  \\
  	SimpleVQA \cite{sun2022deep} &  $448\times448$, $1$ fps &$224\times224$, None& ResNet-50 & SlowFast & MLP & Fix TA, Train SA\&QR &  L1\&Ranking \\
  	FAST-VQA \cite{wu2022fast} &  ---& $224\times224$, $1.6$-$6.4$ fps& --- & Video Swin-T & MLP & End-to-End Training &  PLCC\&Ranking  \\
  	DOVER \cite{wu2022disentangling} &  $224\times224$, $1.6$-$6.4$ fps& $224\times224$, $1.6$-$6.4$ fps& ConvNeXt-T & Video Swin-T & MLP & End-to-End Training &  PLCC\&Ranking  \\
   \hline
  	Ours (Model (\rom{1})) &  $448\times448$, $1$ fps &$224\times224$, $16$ fps& ResNet-50 & SlowFast & MLP & Fix TA, Train SA\&QR &  PLCC \\
    \bottomrule[.15em]
	\end{tabular}
	}
\end{table*}

\subsection{BVQA Models}
\label{sec:bvqa_models}
BVQA models can be loosely categorized into knowledge-driven and data-driven methods.

\noindent\textbf{Knowledge-driven BVQA models} rely heavily on handcrafted features such as natural scene statistics (NSS). A na\"{i}ve BVQA method computes the quality of each frame via popular NSS-based BIQA methods (\eg, NIQE \cite{mittal2012making} and BRISQUE \cite{mittal2012no}), followed by temporal quality pooling \cite{tu2020comparative}. Since motion analysis is regarded as crucial in BVQA to quantify (spatio)temporal distortions, many methods incorporate temporal features into BIQA solutions. V-BLIINDS \cite{saad2014blind} examines spatial naturalness (by NIQE), spatiotemporal statistical regularity (by 2D DCT analysis over frame differences),  and motion coherency (through computed motion vectors). Mittal \textit{et al.} \cite{mittal2015completely} proposed a training-free BVQA model named VIIDEO that exploits intrinsic statistical regularities of natural videos. Korhonen \cite{korhonen2019two} extracted a larger set of spatiotemporal features related to motion regularities, flickering, jerkiness, blur, blockiness, noise, and color, at two scales. Tu~\etal~\cite{tu2021ugc} combined the features from multiple BIQA and BVQA methods, and trained a support vector regressor for quality assessment.

It is believed that understanding semantic video information is beneficial for BVQA. Hence, some studies \cite{tu2021rapique, korhonen2020blind} tried to combine low-level handcrafted features with high-level deep features to boost the BVQA performance. 
 CNN-TLVQM \cite{korhonen2020blind} combines the handcrafted temporal features from TLVQM~\cite{korhonen2019two} and spatial features extracted by the pre-trained ResNet-50 for BIQA. 
RAPIQUE \cite{tu2021rapique} utilizes the handcrafted  NSS and deep features by ResNet-50 pre-trained for object recognition. Despite the long pursuit of
 NSS for BVQA, they exhibit subpar performance, some of which induce high computational complexity.

\noindent\textbf{Data-driven BVQA models} differ mainly in the instantiations of the four building blocks introduced in Section~\ref{sec:introduction} and the optimization strategies. For the \textit{video preprocessor},  a considerable body of work chooses to perform spatiotemporal downsampling to reduce the computational burdens.
For the \textit{spatial quality analyzer}, the prevailing model architectures are  ResNets~\cite{he2016deep} and VGG networks~\cite{simonyan2014very} pre-trained for object recognition~\cite{deng2009imagenet} or  IQA~\cite{ying2020patches,zhang2021uncertainty}. Other advanced architectures such as EfficientNets~\cite{tan2019efficientnet}, ConvNeXts~\cite{liu2022convnet}, vision Transformers (ViTs)~\cite{dosovitskiy2020image} have also been explored~\cite{wang2021rich,wu2022fast,wu2022disentangling}. For the \textit{temporal quality analyzer}, the dominant model architectures are heavily influenced by those in the action recognition field, including 3D ResNets and D3D~\cite{stroud2020d3d}, SlowFast networks~\cite{feichtenhofer2019slowfast}, and video Swin Transformers~\cite{liu2022video}. For the \textit{quality regressor}, priority is given to long-term dependency modeling in the temporal domain. Representative regressors include the simple average pooling, the multilayer perceptron (MLP), the gated recurrent unit (GRU), and deep models for time series (\eg, InceptionTime~\cite{ismail2020inceptiontime}).

With respect to optimization strategies, early methods prefer to fix spatial and temporal quality analyzers, and optimize the quality regressor solely, such as VSFA \cite{li2019quality}, PatchVQ~\cite{ying2021patch}, CoINVQ~\cite{wang2021rich}, and Li22~\cite{li2022blindly}. 
BVQA methods based on pre-trained spatial and temporal quality analyzers benefit from other visual tasks, and mitigate the reliance on human-rated video data for end-to-end training. However,  domain shifts commonly exist between the pretext and BVQA tasks, which are bound to give sub-optimal solutions.

End-to-end optimization of BVQA models is no easy task due to the huge demand for human-rated data, the mass computer memory consumption, and the long training time to convergence. However, if enabled, it gives BVQA models the opportunity to learn quality-aware features from raw video pixels for improved performance. Liu \textit{et al.} \cite{liu2018end} proposed a lightweight BVQA model, V-MEON, by jointly optimizing a 3D-DNN for quality assessment and compression distortion classification. Yi \textit{et al.} \cite{yi2021attention} relied only on the spatial quality analyzer (\ie, a non-local VGG network for attention modeling), leaving temporal modeling to the quality regressor. Wu \textit{et al.} \cite{wu2022fast} extracted fragments (\ie, the spatially spliced and temporally aligned patches)  as input to a video Swin Transformer~\cite{liu2022video}. They \cite{wu2022disentangling} later used the same trick to enable end-to-end optimization, and captured video distortions at the signal and aesthetics levels. As the preliminary work~\cite{sun2022deep}, we developed a simple BVQA method consisting of a multi-scale spatial feature extractor and a pre-trained fixed motion extractor.

To facilitate the comparison of DNN-based BVQA models, we summarize their computational structures as well as optimization strategies in Table \ref{tab:bvqa_models}. It is clear that all BVQA models can be decomposed into the four basic building blocks. Here we limit our BVQA model capacity with minimalistic instantiations of these building blocks to probe the easy dataset problem of existing VQA datasets.

\section{Design of Minimalistic BVQA Models}
\label{proposed_model}

In this section, we design a family of minimalistic BVQA models with the goal of analyzing existing VQA datasets. Our models consist of four basic building blocks: a video preprocessor, a spatial quality analyzer, an optional temporal quality analyzer, and a quality regressor.

\subsection{Formulation of the BVQA Problem}
\label{model_design}
We first formulate the problem of  BVQA. Assume a video $\bm{x}= \{\bm x_i\}_{i=0}^{N-1}$, where $\bm x_i\in\mathbb{R}^{H\times W \times 3}$ denotes the $i$-th frame. $H$ and $W$ are the height and width of each frame, and $N$ is the total number of frames. The objective of a BVQA model $q_{\bm{w}}(\cdot):\mathbb{R}^{H\times W \times 3\times N} \mapsto \mathbb{R}$, parameterized by a vector $\bm w$, is to compute a scalar as an approximation to the true perceptual quality, $q(\bm x)$:
\begin{align}
    \hat{q} = q_{\bm w}(\bm x).
\end{align}
We may conveniently break down $q_{\bm w}(\cdot)$ into four basic building blocks. The first is a video preprocessor $\bm{p} (\cdot)$ to spatiotemporally downsample $\bm x$, leading to a set of $K$ key frames $\bm z = [\bm z_0, \bm z_1, \ldots, \bm z_{K-1}]$.  Each key frame $\bm z_i$ may be optionally accompanied by a video chunk $\bm v^{(i)}$ for temporal quality analysis:
\begin{align}
\bm{z}, \mathcal{V} = \bm{p}(\bm{x}),
\end{align}
where $\mathcal{V} = \{\bm v^{(i)}\}_{i=0}^{K-1}$.
The second is a spatial quality analyzer $\bm{s}(\cdot)$ that takes a key frame as input and extracts quality-aware spatial features:
\begin{align}
    \bm{s}_i = \bm{s}(\bm z_i), \quad \mathrm{for}\quad i = 0, 1,\ldots, K-1.
\end{align}
The third is an optional temporal quality analyzer $\bm{t}(\cdot)$ to extract temporal features from $\bm v^{(i)}$:
\begin{align}
    \bm{t}_i = \bm{t}(\bm v^{(i)}), \quad \mathrm{for}\quad i = 0, 1,\ldots, K-1.
\end{align}
The last is a quality regressor $a(\cdot)$ that maps the spatial and temporal quality features into a global quality score:
\begin{align}
\hat{q}  = a(\bm s,\bm t),
\end{align}
where $\bm s = [\bm s_0, \bm s_1, \ldots, \bm s_{K-1}]$ and $\bm t = [\bm t_0, \bm t_1, \ldots, \bm t_{K-1}]$.
\subsection{Video Preprocessor}
Though it is convenient to record videos in 4K resolution and $60$ fps, inputting such a raw video to the BVQA model would induce extremely high computation \cite{li2019quality,li2022blindly}. It also poses a grand challenge to end-to-end optimization of BVQA models. Being minimalistic, we perform aggressive spatiotemporal downsampling as video preprocessing. We first temporally downsample the raw video $\bm x=[\bm x_0,\bm x_1,
\ldots,\bm x_{N-1}]$ from $R$ to $R_a$ fps:
\begin{align}
\bm y_{i} = \bm{x}_{\lfloor R/R_a\times (0.5 + i) \rfloor},
\end{align}
where $\bm{y}_i$ is the $i$-th key frame subject to spatial downsampling, $\lfloor\cdot\rfloor$ is the floor function, and $K= \lfloor NR_a/R\rfloor$. We then spatially downsample each key frame $\bm y_i$ to $\bm z_i$ using the bilinear kernel while keeping the aspect ratio, such that the shorter side has a length of $L_s$.

\begin{figure}[!t]
	\centering
	\includegraphics[height=2.4in]{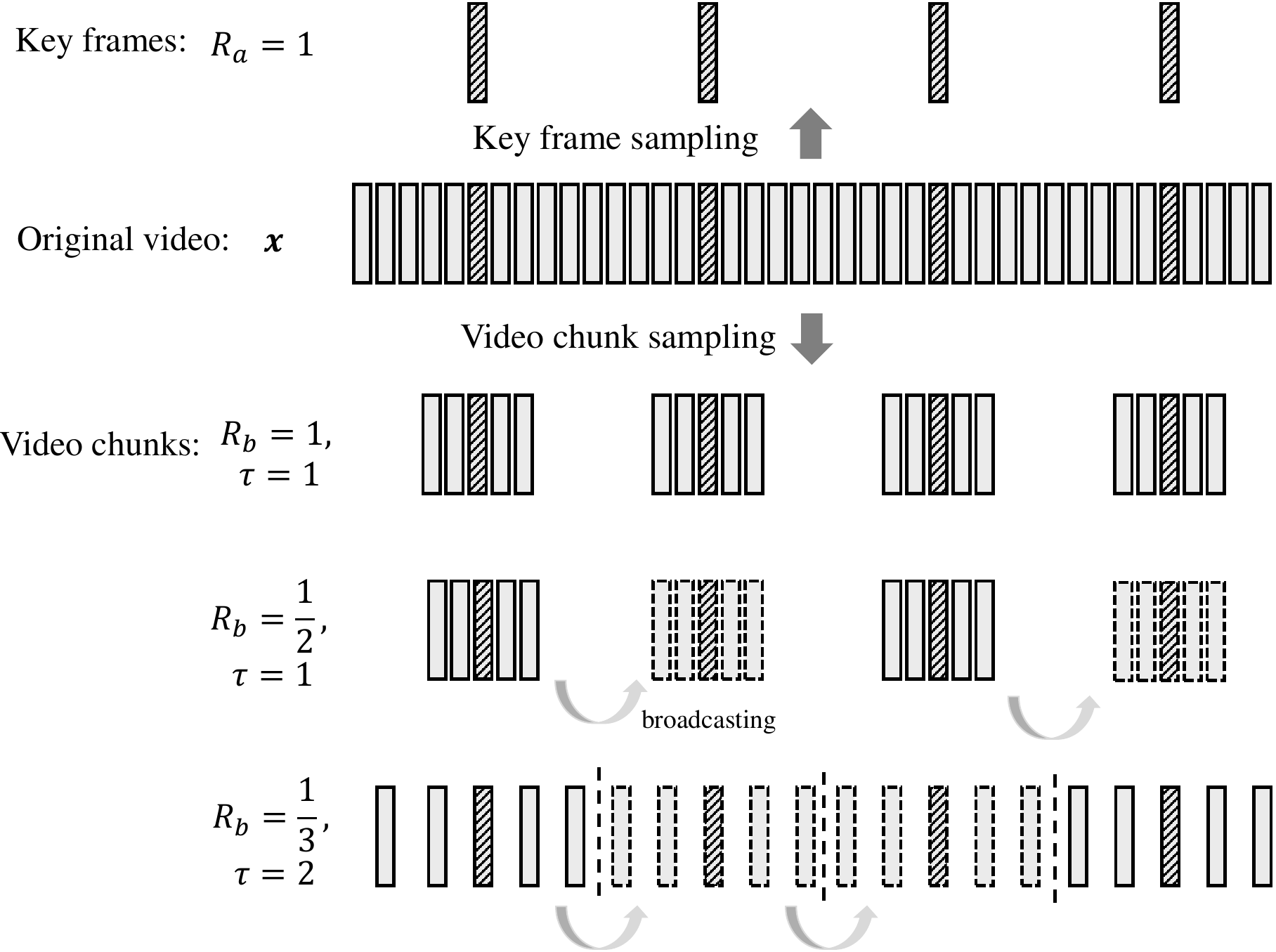}
	\caption{Illustration of the video preprocessor. In this example, the frame number, $N$, and the frame rate, $R$, of the video $\bm{x}$ are $40$ and $10$ fps, respectively. The frame number $T$ of each video chunk is $5$. For brevity, we omit the procedure of spatial downsampling.}
	\label{fig:video_preprocessor}
\end{figure}

Optionally, for each key frame $\bm y_i$, we sample a video chunk of $T+1$ frames $\bm{y}^{(i)}= [\bm y_0^{(i)},\bm y_2^{(i)},\ldots,\bm y_{T}^{(i)}]$
 from $\bm x$ at a frame interval $\tau$ for temporal quality analysis:
\begin{align}\label{eq:vc}
\bm{y}_j^{(i)} = \bm{x}_{\lfloor R/R_a(0.5+i) \rfloor +\tau(j-\lfloor T/2\rfloor)},
\end{align}
where $\bm y_i = \bm y^{(i)}_{\lfloor T/2\rfloor }$ is centered at the video chunk. 
As  temporal features are relatively stable under different spatial resolutions~\cite{feichtenhofer2019slowfast}, we resize $\bm y^{(i)}$ to $\bm{v}^{(i)}$ with resolution of $L_t\times L_t$, where  $L_t< L_s$. To downgrade the importance of the temporal features, we may further temporally downsample the key frames $\bm y=[\bm y_0,\bm y_1,
\ldots,\bm y_{K-1}]$ from $R_a$ to $R_b$ fps before video chunk extraction. In this case, the missing video chunks for decimated key frames can be na\"{i}vely broadcasted from the computed video chunks, as shown in Fig.~\ref{fig:video_preprocessor}.

\subsection{Spatial Quality Analyzer}
The spatial quality analyzer $\bm{s}(\cdot)$ takes one key frame as input, and extracts quality-aware spatial features. It is computationally close to BIQA models, and many previous BVQA models \cite{ying2021patch,li2022blindly} directly adopt BIQA models for spatial quality analysis. Nevertheless, the impact of spatial distortions in videos may differ from that in images. For example, a fast-moving object may appear blurred in a single frame, but looks good when viewed in the video. In this case, a BIQA model may over-penalize spatial distortions, causing inaccurate estimation of perceptual video quality.  We take into account these subtle differences by training DNN-based BIQA models with different quality-aware initializations as the spatial quality analyzers. We choose two popular backbones,  ResNet-50 \cite{he2016deep} and Swin Transformer-B \cite{liu2021swin}, and strip their classification heads. Being minimalistic,  we do not employ advanced feature extraction strategies, such as multi-stage feature fusion \cite{sun2022deep} and global average and standard deviation pooling  \cite{li2019quality,li2022blindly,sun2022deep}, and only apply global average pooling to the last-stage feature maps. 

As stated in Section \ref{sec:bvqa_models}, leveraging quality-aware initializations has been a common practice for DNN-based BVQA. For the spatial quality analyzer, we explore initializations 1) pre-trained on ImageNet, 2) pre-trained on IQA datasets, and 3) pre-trained on  VQA datasets (\eg, pre-trained on LSVQ and tested on LIVE-VQC).
Other initializations (\eg, resulting from unsupervised learning~\cite{he2020momentum,he2022masked,radford2021learning}) will be ablated in Section \ref{subsubsec:oi}. 

\subsection{Temporal Quality Analyzer} 
We allow an optional temporal quality analyzer $\bm t(\cdot)$, and try to make its design as simple as possible.  $\bm t(\cdot)$ independently extracts the temporal features from the $i$-th video chunk $\bm v^{(i)}$ centered at the $i$-th key frame $\bm z_i$ as a form of short-term memory. Being minimalistic, temporal dependencies across different video chunks as the long-term memory is not modeled. We also choose to freeze the temporal quality analyzer,   making it more of a local motion analyzer. 
We instantiate the temporal quality analyzer by the temporal pathway of the SlowFast network~\cite{feichtenhofer2019slowfast}, pre-trained on the Kinetics-400 dataset~\cite{carreira2017quo}. Similar to the spatial feature extraction, we throw away the original classification head, and compute the temporal features by global average pooling of the last-stage feature maps.

\subsection{Quality Regressor}
The last step is to map the extracted spatial (and temporal) features to a single scalar as the global quality score. In previous studies, sequential models such as GRU \cite{li2019quality} and InceptionTime \cite{ying2021patch} are adopted to capture useful long-term dependencies among key frames (and the associated video chunks). Being minimalistic, we intentionally opt for a much simpler fully-connected (FC) layer (as a special case of the MLP), regressing the $i$-th local quality score $\hat{q}_i$ based on the concatenated spatial and temporal features:
\begin{align}
\hat{q}_i = [\bm{s}_i{^\intercal}, \bm{t}_i{^\intercal}] \bm{\beta}, \quad \mathrm{for}\quad i = 0, 1, \ldots, K-1,
\end{align}
where $\bm{\beta}$ denotes the learnable linear projection vector.
Finally, we aggregate frame-level (or chunk-level) quality estimates into the video-level quality score by simple averaging:
\begin{align}\label{eq:gsa}
\hat{q} = \frac{1}{K}\sum_{i=0}^{K-1} \hat{q}_i.
\end{align}

\subsection{Loss function}
\label{loss_function}
The remaining is to specify a proper loss function for end-to-end BVQA model optimization. Conventionally, BVQA is regarded as a regression task, which can be supervised by 
the mean absolute error (MAE)  and the mean squared error (MSE) \cite{li2019quality, ying2021patch, korhonen2020blind, liu2018end, wang2021rich, yi2021attention} to predict \textit{absolute} quality (\ie, MOSs). Alternatively, we may look at \textit{relative} quality prediction by formulating BVQA as a learning-to-rank problem. Thus, pairwise ranking losses such as the fidelity loss under the Thurstone-Mosteller models~\cite{thurstone1927law} (or the Bradley-Terry models~\cite{bradley1952rank}) and listwise ranking losses such as the cross entropy loss under the permutation probabilities, 
differentiable Spearman's rank correlation coefficient (SRCC) \cite{li2022blindly, wu2022disentangling}, and PLCC \cite{li2022blindly, wu2022fast,wu2022disentangling} are viable choices.
Here, we adopt PLCC as the sole loss for optimization of our BVQA models.

\section{Computational Analysis of VQA Datasets}
\label{experimental_results}

In this section, we first present in detail the experimental settings for comparison of our minimalistic  BVQA models against seven representative methods with different design philosophies. Based on the experimental results, we then conduct a computational analysis of eight VQA datasets, pointing out the overlooked easy dataset problem.

\subsection{Experimental Settings}
\label{experimental_settings}
\noindent\textbf{VQA Datasets}.
We evaluate  BVQA models on eight  VQA datasets introduced in Section \ref{vqa_dataset} and summarized in Table \ref{overview_vqa_dataset}. These include CVD2014 \cite{nuutinen2016cvd2014}, LIVE-Qualcomm \cite{ghadiyaram2017capture}, KoNViD-1k \cite{hosu2017konstanz}, LIVE-VQC \cite{sinno2018large}, YouTube-UGC \cite{wang2019youtube}, LIVE-YT-Gaming \cite{yu2022subjective}, LBVD \cite{chen2019qoe}, and LSVQ \cite{ying2021patch}. For the LSVQ dataset, we follow the original dataset splitting \cite{ying2021patch}, in which the training set consists of $28,013$ videos, and the default and 1080p test sets contain $7,220$ and $3,561$ videos, respectively. We randomly sample $20\%$ videos from the training set for validation, and report the results on the two test sets. For the remaining seven datasets, we randomly split each into the training, validation, and test sets with a ratio of $6:2:2$  for $10$ times, and report the median results.

\noindent\textbf{Competing Methods.} We select seven representative BVQA models including three knowledge-driven methods, BRISQUE~\cite{mittal2012no}, TLVQM~\cite{korhonen2019two}, and RAPIQUE~\cite{tu2021rapique}, and four data-driven methods, VSFA~\cite{li2019quality}, PatchVQ~\cite{ying2021patch}, Li22~\cite{li2022blindly}, and FAST-VQA~\cite{wu2022fast} for comparison. BRISQUE is a popular NSS-based BIQA model, for which we calculate the spatial features at  $1$ fps. TLVQM and RAPIQUE extract a set of spatiotemporal features for quality evaluation. 
VSFA and Li22 rely on pre-trained models to extract spatiotemporal features, and only train the quality regressor. FAST-VQA is a state-of-the-art BVQA model that end-to-end optimizes a variant of video Swin Transformer~\cite{liu2022video}. For a fair comparison, we retrain all competing models on seven small-scale VQA datasets based on our dataset splitting strategy, following their original optimization procedures. For the large-scale LSVQ with held-out test sets, we copy the results from the respective papers.

We instantiate ten models from our minimalistic BVQA model family: \textbf{(\rom{1})} ResNet-50 pre-trained on ImageNet-1K as the spatial quality analyzer with no temporal quality analyzer (as the baseline model); \textbf{(\rom{2})} pre-training (\rom{1}) on the combination of four IQA datasets\footnote{These are BID \cite{ciancio2011no}, LIVE Challenge \cite{ghadiyaram2015massive}, KonIQ-10k \cite{hosu2020koniq}, and SPAQ \cite{fang2020perceptual}. The enabling training strategy was proposed in~\cite{zhang2021uncertainty}.}; \textbf{(\rom{3})} pre-training (\rom{1}) on the training set of LSVQ\footnote{This variant shall be tested on the other seven VQA datasets.}; \textbf{(\rom{4})} adding the temporal analyzer to (\rom{1}); \textbf{(\rom{5})} adding the temporal analyzer to (\rom{2}); \textbf{(\rom{6})} adding the temporal analyzer to (\rom{3}); \textbf{(\rom{7})} to \textbf{(\rom{10})}: replacing ResNet-50 with Swin Transformer-B as the spatial quality analyzer in (\rom{1}), (\rom{3}), (\rom{4}), and (\rom{6}), respectively.

\noindent\textbf{Evaluation Criteria.}
We adopt two criteria to evaluate the performance of VQA models:  SRCC and PLCC, which indicate the prediction monotonicity and prediction linearity, respectively. Before calculating PLCC, we follow the procedure in \cite{antkowiak2000final} to map model predictions to  MOSs using a monotonic four-parameter logistic function to compensate for prediction nonlinearity.

\noindent\textbf{Implementation Details}.
For the video preprocessor, we extract key frames at $1$ fps (\eg, $K=10$ for a $10$-second video clip) and set the shorter side of the key frames, $L_s$, to $448$ for ResNet-50 and $384$ for Swin Transformer-B, and the spatial resolution of the video chunks, $L_t$, to $224$, respectively. The temporal sampling rates of the key frames, $R_a$, and the video chunks, $R_b$, are set to $1$ and $0.5$ fps, respectively. We sample $T=32$ frames for each video chunk at a frame interval $\tau = 1$. We choose two backbone networks: ResNet-50 \cite{he2016deep} and Swin Transformer-B \cite{liu2021swin} pre-trained on ImageNet-1k \cite{deng2009imagenet} as the spatial quality analyzers, and use the fast pathway of SlowFast Network (ResNet-50)~\cite{feichtenhofer2019slowfast} pre-trained on Kinetics-400~\cite{carreira2017quo} as the temporal quality analyzer.

Training is carried out by optimizing PLCC using the Adam method~\cite{kingma2014adam} with an initial learning rate of $10^{-5}$ and a minibatch size of $8$. In the training stage, the input frames are randomly cropped to $448\times448$ for ResNet-50 and $384\times384$ for Swin Transformer-B. In the validation and test stages, the input frames are center-cropped to the same spatial resolutions as in the training stage. We decay the learning rate by a factor of $10$ after $2$ epochs for LSVQ, and after $10$ epochs for other datasets. Depending on the dataset size, the maximum number of epochs is set to $10$ for LSVQ, $30$ for KoNViD-1k, YouTube-UGC, and LBVD, and $50$ for CVD2014, LIVE-Qualcomm, LIVE-VQC, and LIVE-YT-Gaming, respectively.

\begin{table*}
\centering
\renewcommand{\arraystretch}{1.25}
\caption{Performance comparison in terms of SRCC and PLCC of the proposed minimalistic BVQA models against seven competing methods on eight VQA datasets. Li22 relies on model parameters pre-trained on the combination of four IQA datasets, while the results of FAST-VQA on seven small-scale VQA datasets (except for LSVQ) are based on  model parameters pre-trained on LSVQ. W.A. means the weighted average result, with the weighting proportional to the number of videos in each dataset. The top-2 results on each dataset are highlighted in bold}
\label{performance_proposed_method}
\resizebox{1\textwidth}{!}{
\begin{tabular}{ll|cc|cc|cc|cc|cc}
\toprule[.15em]
 \multirow{2}{*}{Method} & \multirow{2}{*}{}  & \multicolumn{2}{c|}{CVD2014} & \multicolumn{2}{c|}{LIVE-Qualcomm} & \multicolumn{2}{c|}{KoNViD-1k} & \multicolumn{2}{c|}{LIVE-VQC}& \multicolumn{2}{c}{YouTube-UGC}\\
  &  & SRCC  & PLCC & SRCC & PLCC & SRCC & PLCC & SRCC & PLCC & SRCC & PLCC\\

\hline
\rowcolor{tablegray}
\multicolumn{12}{l}{\textit{Competing Methods as Reference}} \\
BRISQUE~\cite{mittal2012no} &  &  0.619&  0.702  &  0.463 & 0.553 & 0.672  & 0.673 &  0.598 & 0.626 & 0.335  &  0.379\\
TLVQM~\cite{korhonen2019two} &  & 0.720 & 0.756   &  0.747 & 0.781 & 0.770  & 0.761 & 0.747  & 0.781 &  0.735 & 0.717 \\
RAPIQUE~\cite{tu2021rapique} &  & 0.772 & 0.792   & 0.636  & 0.673 & 0.811  & 0.819 & 0.740  & 0.764 & 0.774  &  0.781\\
VSFA~\cite{li2019quality} &  & 0.850 &  0.869  & 0.708  & 0.774 &  0.794 & 0.799 &  0.718 & 0.771 & 0.787  & 0.789 \\
PatchVQ~\cite{ying2021patch} &  & --- &  ---  &  --- &  ---&  0.791 & 0.786 & 0.827  & 0.837 &  --- & --- \\
Li22~\cite{li2022blindly} &  & 0.863&  0.883 &  \textbf{0.833} & \textbf{0.837} & 0.839  & 0.830 & 0.841  & 0.839&   0.825 &  0.818 \\
FAST-VQA~\cite{wu2022fast} &  &  \textbf{0.883} &  \textbf{0.901}  &  0.807  &   0.814 &   \textbf{0.893} &  \textbf{0.887} &    \textbf{0.853} &   \textbf{0.873} &  0.863 & 0.859 \\

\hline
\rowcolor{tablegray}
\multicolumn{12}{l}{\textit{Proposed Methods}} \\
\multicolumn{2}{l|}{(\rom{1}): ResNet-50 as baseline} &0.867 &   0.889 & 0.651  & 0.718 & 0.764   & 0.781  &  0.653 &  0.719&  0.822 & 0.826 \\ 
\cdashline{1-12}
\multicolumn{2}{l|}{(\rom{2}): \textit{(\rom{1}) + IQA pre-trained}}& 0.866&   0.882 &  0.762 &  0.805&  0.822 & 0.830 & 0.736  & 0.788 &  0.847 &  0.843 \\
\multicolumn{2}{l|}{\textit{\textcolor{gray}{Improvement over baseline}}} &\textcolor{gray}{-0.1\%} &  \textcolor{gray}{-0.8\%} &  \textcolor{gray}{+17.1\%} & \textcolor{gray}{+12.1\%} &  \textcolor{gray}{+7.6\%} & \textcolor{gray}{+6.3\%} & \textcolor{gray}{+12.7\%}  & \textcolor{gray}{+9.6\%}& \textcolor{gray}{+3.0\%}  & \textcolor{gray}{+2.1\%} \\
\multicolumn{2}{l|}{(\rom{3}): \textit{(\rom{1}) + VQA pre-trained}}&0.860 & 0.889  &  0.774 & 0.800 &  0.861 &  0.861&  0.722 &0.761 &  0.834 & 0.843 \\
\multicolumn{2}{l|}{\textit{\textcolor{gray}{Improvement over baseline}}} &\textcolor{gray}{-0.8\%} &  \textcolor{gray}{0.0\%} &  \textcolor{gray}{+18.9\%} & \textcolor{gray}{+11.4\%} &  \textcolor{gray}{+12.7\%} & \textcolor{gray}{+10.2\%} & \textcolor{gray}{+10.6\%}  & \textcolor{gray}{+5.8\%}& \textcolor{gray}{+1.5\%}  & \textcolor{gray}{+2.1\%} \\
\multicolumn{2}{l|}{(\rom{4}): \textit{(\rom{1}) + motion features}} &0.852 &   0.871 &  0.693 &  0.732 &  0.797 & 0.803 & 0.752   &  0.798 & 0.810   &  0.809  \\
\multicolumn{2}{l|}{\textit{\textcolor{gray}{Improvement over baseline}}} &\textcolor{gray}{-1.7\%} &  \textcolor{gray}{-2.0\%} &  \textcolor{gray}{+6.5\%} & \textcolor{gray}{+1.9\%} &  \textcolor{gray}{+4.4\%} & \textcolor{gray}{+2.8\%} & \textcolor{gray}{+15.2\%}  & \textcolor{gray}{+11.0\%}& \textcolor{gray}{-1.5\%}  & \textcolor{gray}{-2.1\%} \\
\multicolumn{2}{l|}{(\rom{5}): \textit{(\rom{2}) + motion features}}& 0.879&  \textbf{0.899} & 0.749  & 0.805 &  0.843 & 0.849 &  0.795 & 0.819&  0.830 & 0.835 \\
\multicolumn{2}{l|}{\textit{\textcolor{gray}{Improvement over baseline}}} &\textcolor{gray}{+1.0\%} &  \textcolor{gray}{+1.4\%} &  \textcolor{gray}{+15.1\%} & \textcolor{gray}{+12.1\%} &  \textcolor{gray}{+10.3\%} & \textcolor{gray}{+8.7\%} & \textcolor{gray}{+21.7\%}  & \textcolor{gray}{+13.9\%}& \textcolor{gray}{+1.0\%}  & \textcolor{gray}{+1.1\%} \\
\multicolumn{2}{l|}{(\rom{6}): \textit{(\rom{3}) + motion features}}&0.876 & 0.893  &  0.753 &  0.800 &  0.859 & 0.864 & 0.813  & 0.843& 0.845  &  0.853 \\
\multicolumn{2}{l|}{\textit{\textcolor{gray}{Improvement over baseline}}} &\textcolor{gray}{+1.0\%} &  \textcolor{gray}{+0.5\%} &  \textcolor{gray}{+15.7\%} & \textcolor{gray}{+11.4\%} &  \textcolor{gray}{+12.5\%} & \textcolor{gray}{+10.6\%} & \textcolor{gray}{+24.5\%}  & \textcolor{gray}{+17.2\%}& \textcolor{gray}{+2.8\%}  & \textcolor{gray}{+3.3\%} \\
\hline

(\rom{7}): Swin Transformer-B as baseline &  & 0.833&  0.860  &  0.710 &  0.711 & 0.823   & 0.830  &  0.683 &  0.735 &  0.832 & 0.837 \\
\cdashline{1-12}
\multicolumn{2}{l|}{(\rom{8}): \textit{(\rom{7}) + VQA pre-trained}}& 0.862 &  0.888 &  0.768 & 0.808 & 0.850  & 0.857 &  0.765 & 0.821& \textbf{0.877}  &  \textbf{0.873} \\
\multicolumn{2}{l|}{\textit{\textcolor{gray}{Improvement over baseline}}} &\textcolor{gray}{+3.5\%} &  \textcolor{gray}{+3.3\%} &  \textcolor{gray}{+8.2\%} & \textcolor{gray}{+13.6\%} &  \textcolor{gray}{+3.3\%} & \textcolor{gray}{+3.2\%} & \textcolor{gray}{+12.0\%}  & \textcolor{gray}{+11.7\%}& \textcolor{gray}{+5.4\%}  & \textcolor{gray}{+4.3\%} \\

\multicolumn{2}{l|}{(\rom{9}): \textit{(\rom{7}) + motion features}}&0.845 &   0.869 & 0.728  & 0.748 &  0.851 & 0.854 &  0.772 & 0.809& 0.835  &  0.842  \\
\multicolumn{2}{l|}{\textit{\textcolor{gray}{Improvement over baseline}}} &\textcolor{gray}{+1.4\%} &  \textcolor{gray}{+1.0\%} &  \textcolor{gray}{+2.5\%} & \textcolor{gray}{+5.2\%} &  \textcolor{gray}{+3.4\%} & \textcolor{gray}{+2.9\%} & \textcolor{gray}{+13.0\%}  & \textcolor{gray}{+10.1\%}& \textcolor{gray}{+0.4\%}  & \textcolor{gray}{+0.6\%} \\

\multicolumn{2}{l|}{(\rom{10}): \textit{(\rom{8}) + motion features}}& \textbf{0.882} &   \textbf{0.899} &  \textbf{0.835} & \textbf{0.837} & \textbf{0.889}  & \textbf{0.890} &  \textbf{0.842} & \textbf{0.854}&  \textbf{0.890} &  \textbf{0.891} \\
\multicolumn{2}{l|}{\textit{\textcolor{gray}{Improvement over baseline}}} &\textcolor{gray}{+5.9\%} &  \textcolor{gray}{+4.5\%} &  \textcolor{gray}{+17.6\%} & \textcolor{gray}{+17.7\%} &  \textcolor{gray}{+8.0\%} & \textcolor{gray}{+7.2\%} & \textcolor{gray}{+23.3\%}  & \textcolor{gray}{+16.2\%}& \textcolor{gray}{+7.0\%}  & \textcolor{gray}{+6.5\%} \\

\bottomrule[.15em]
 \multirow{2}{*}{Method} & \multirow{2}{*}{}  & \multicolumn{2}{c|}{LBVD} & \multicolumn{2}{c|}{LSVQ} & \multicolumn{2}{c|}{LSVQ-1080p} & \multicolumn{2}{c|}{LIVE-YT-Gaming}& \multicolumn{2}{c}{W.A.}\\
  &  & SRCC  & PLCC & SRCC & PLCC & SRCC & PLCC & SRCC & PLCC & SRCC & PLCC\\

\hline
\rowcolor{tablegray}
\multicolumn{12}{l}{\textit{Competing Methods as Reference}} \\
BRISQUE \cite{mittal2012no} &  & 0.477 &  0.510  &  0.579 & 0.576 &  0.497 & 0.531 &  0.564 & 0.572 & 0.543  &  0.558\\
TLVQM \cite{korhonen2019two} &  & 0.794 &  0.804  & 0.772  & 0.774 & 0.589  & 0.616 &  0.740 & 0.775 & 0.726  & 0.735 \\
RAPIQUE \cite{tu2021rapique} &  & 0.781 &  0.774  &  0.827 & 0.819 &  0.655 & 0.701 & 0.771  & 0.815 & 0.771  & 0.782 \\
VSFA \cite{li2019quality} &  & 0.834 &   0.830 &  0.801 & 0.796 &  0.675 &  0.704 &  0.784 &  0.819 & 0.769&0.778\\
PatchVQ \cite{ying2021patch} &  & --- &  ---  & 0.827  & 0.828 & 0.711  & 0.739 &  --- &  ---&  --- &---  \\
Li22 \cite{li2022blindly} &  &\textbf{0.891} &  \textbf{0.887} & 0.852 & 0.854 &  0.772 & 0.788 &   0.852 & 0.868 & 0.833&0.837\\
FAST-VQA \cite{wu2022fast} &  & 0.804 &  0.809  &  {0.876} & {0.877} & {0.779}  & {0.814} & \textbf{0.869}  &  0.880  &\textbf{0.848} &\textbf{0.857}\\

\hline
\rowcolor{tablegray}
\multicolumn{12}{l}{\textit{Porposed  Methods}} \\
\multicolumn{2}{l|}{(\rom{1}): ResNet-50 as baseline}  & 0.704&   0.706 &  0.833 & 0.833 &  0.711 & 0.761&  0.800 &   0.853& 0.784&0.802\\
\cdashline{1-12}
\multicolumn{2}{l|}{(\rom{2}): \textit{(\rom{1}) + IQA pre-trained}}&0.716 &  0.718  & 0.834  &  0.833&  0.713 &  0.765 &  0.835 & 0.877 &  0.795 &  0.811\\
\multicolumn{2}{l|}{\textit{\textcolor{gray}{Improvement over baseline}}} &\textcolor{gray}{+1.7\%} &  \textcolor{gray}{+1.7\%} &  \textcolor{gray}{+0.1\%} & \textcolor{gray}{+0.0\%} &  \textcolor{gray}{+0.3\%} & \textcolor{gray}{+0.5\%} & \textcolor{gray}{+4.4\%}  & \textcolor{gray}{+2.8\%}& \textcolor{gray}{+1.4\%}  & \textcolor{gray}{+1.1\%} \\
\multicolumn{2}{l|}{(\rom{3}): \textit{(\rom{1}) + VQA pre-trained}}&0.705 & 0.705  & 0.833  &  0.833 & 0.711  &  0.761 & 0.822  & 0.868&  0.795 &  0.810\\
\multicolumn{2}{l|}{\textit{\textcolor{gray}{Improvement over baseline}}} &\textcolor{gray}{+0.1\%} &  \textcolor{gray}{-0.1\%} &  \textcolor{gray}{+0.0\%} & \textcolor{gray}{+0.0\%} &  \textcolor{gray}{+0.0\%} & \textcolor{gray}{+0.0\%} & \textcolor{gray}{+2.7\%}  & \textcolor{gray}{+1.8\%}& \textcolor{gray}{+1.4\%}  & \textcolor{gray}{+1.0\%} \\

\multicolumn{2}{l|}{(\rom{4}): \textit{(\rom{1}) + motion features}}& 0.882 &  0.881   &  0.864  &  0.862  & 0.748   & 0.789  &  0.826  & 0.857    & 0.822 &0.834 \\
\multicolumn{2}{l|}{\textit{\textcolor{gray}{Improvement over baseline}}} &\textcolor{gray}{+25.3\%} &  \textcolor{gray}{+24.8\%} &  \textcolor{gray}{+3.7\%} & \textcolor{gray}{+3.5\%} &  \textcolor{gray}{+5.2\%} & \textcolor{gray}{+3.7\%} & \textcolor{gray}{+3.2\%}  & \textcolor{gray}{+0.5\%}& \textcolor{gray}{+4.8\%}  & \textcolor{gray}{+4.0\%} \\

\multicolumn{2}{l|}{(\rom{5}): \textit{(\rom{2}) + motion features}}& \textbf{0.884}&   0.877 & 0.864    &  0.863   & 0.756    &  0.792   & 0.839   &   \textbf{0.883} & 0.825 &0.836 \\
\multicolumn{2}{l|}{\textit{\textcolor{gray}{Improvement over baseline}}} &\textcolor{gray}{+25.6\%} &  \textcolor{gray}{+24.2\%} &  \textcolor{gray}{+3.7\%} & \textcolor{gray}{+3.6\%} &  \textcolor{gray}{+6.3\%} & \textcolor{gray}{+4.1\%} & \textcolor{gray}{+4.9\%}  & \textcolor{gray}{+3.5\%}& \textcolor{gray}{+5.2\%}  & \textcolor{gray}{+4.2\%} \\

\multicolumn{2}{l|}{(\rom{6}): \textit{(\rom{3}) + motion features}}& 0.878&  0.876 & 0.864  &  0.862 &  0.748 &  0.789 &  0.841 &   0.880 & 0.833 &0.845 \\
\multicolumn{2}{l|}{\textit{\textcolor{gray}{Improvement over baseline}}} &\textcolor{gray}{+24.7\%} &  \textcolor{gray}{+24.1\%} &  \textcolor{gray}{+3.7\%} & \textcolor{gray}{+3.5\%} &  \textcolor{gray}{+5.2\%} & \textcolor{gray}{+3.7\%} & \textcolor{gray}{+5.1\%}  & \textcolor{gray}{+3.2\%}& \textcolor{gray}{+6.2\%}  & \textcolor{gray}{+5.4\%} \\

\hline

(\rom{7}): Swin Transformer-B as baseline &  & 0.692&   0.698 & 0.860  &  0.859 &  0.758 &  0.802 &  0.810 & 0.853 & 0.809&0.824\\
\cdashline{1-12}
\multicolumn{2}{l|}{(\rom{8}): \textit{(\rom{7}) + VQA pre-trained}}& 0.694 &  0.695 &  0.860 & 0.859 &  0.758 & 0.802 &  0.847 & 0.875&  0.821 &  0.835 \\
\multicolumn{2}{l|}{\textit{\textcolor{gray}{Improvement over baseline}}} &\textcolor{gray}{+0.3\%}  &  \textcolor{gray}{-0.4\%}  &  \textcolor{gray}{+0.0\%}  & \textcolor{gray}{+0.0\%}  &  \textcolor{gray}{+0.0\%}  & \textcolor{gray}{+0.0\%}  & \textcolor{gray}{+4.6\%}   & \textcolor{gray}{+2.6\%} & \textcolor{gray}{+1.5\%}  & \textcolor{gray}{+1.3\%} \\

\multicolumn{2}{l|}{(\rom{9}): \textit{(\rom{7}) + motion features}}&0.880  &  \textbf{0.884}   &  \textbf{0.881 } & \textbf{0.879 } &  \textbf{0.781 } & \textbf{0.820 } &  0.845  &  0.870  & 0.845 &\textbf{0.857} \\
\multicolumn{2}{l|}{\textit{\textcolor{gray}{Improvement over baseline}}} &\textcolor{gray}{+27.2\%} &  \textcolor{gray}{+26.6\%} &  \textcolor{gray}{+2.4\%} & \textcolor{gray}{+2.3\%} &  \textcolor{gray}{+3.0\%} & \textcolor{gray}{+2.2\%} & \textcolor{gray}{+4.3\%}  & \textcolor{gray}{+2.0\%}& \textcolor{gray}{+4.4\%}  & \textcolor{gray}{+4.0\%} \\

\multicolumn{2}{l|}{(\rom{10}): \textit{(\rom{8}) + motion features}}& 0.882 &  0.882  & \textbf{0.881 } &  \textbf{0.879 }&  \textbf{0.781 } & \textbf{0.820 } &  \textbf{0.857}  & \textbf{0.888 }  & \textbf{0.857 }&\textbf{0.867 }\\
\multicolumn{2}{l|}{\textit{\textcolor{gray}{Improvement over baseline}}} &\textcolor{gray}{+27.5\%} &  \textcolor{gray}{+26.4\%} &  \textcolor{gray}{+2.4\%} & \textcolor{gray}{+2.3\%} &  \textcolor{gray}{+3.0\%} & \textcolor{gray}{+2.2\%} & \textcolor{gray}{+5.8\%}  & \textcolor{gray}{+4.1\%}& \textcolor{gray}{+5.9\%}  & \textcolor{gray}{+5.2\%} \\

\bottomrule[.15em]
\end{tabular}
}
\end{table*}

\subsection{Main Computational Analysis}
\label{computational_analysis}
\label{sec:performance_analysis}Table~\ref{performance_proposed_method} shows the SRCC and PLCC results of our BVQA models with emphasis on relative improvements, based on which we perform a computational analysis of each dataset in chronological order. The seven competing methods serve as performance reference.

\noindent\textbf{CVD2014}. Model (\rom{1}), closest to a BIQA solution with ResNet-50 as the spatial quality analyzer and no temporal quality analyzer, is able to achieve highly nontrivial results, outperforming all competing models except FAST-VQA~\cite{wu2022fast}. We consider this as a strong indication that CVD2014 suffers from the easy dataset problem due to a poor sample selection strategy. In fact, the videos in CVD2014 were captured using nearly still cameras, with minimal temporal distortions. Meanwhile, the scene diversity was manually optimized, which is bound to be sub-optimal. Our claim is further supported by the results of Models (\rom{4}) to (\rom{6}), in which the incorporation of motion features by the temporal quality analyzer provides little to no improvement. In addition, due to the limited number of videos in CVD2014, increasing the capacity of the spatial quality analyzer from ResNet-50 to Swin Transformer-B leads to overfitting. Model (\rom{8}) that fine-tunes Swin Transformer-B from the VQA pre-trained initialization restores some performance, but is still inferior to Model (\rom{1}).

\noindent\textbf{LIVE-Qualcomm}. Although Model (\rom{1}) lags behind several competing methods, it quickly catches up when starting from IQA and VQA pre-trained initializations. This suggests that despite its simplicity, Model (\rom{1}) has the capacity to handle LIVE-Qualcomm, provided that the adopted optimization technique converges to good local optima. Moreover, by comparing Model (\rom{4}) to Models (\rom{2}) and (\rom{3}) (and Model (\rom{9}) to Model (\rom{8})), it is clear that a better initialization (as part of the optimization technique) for the spatial quality analyzer is more important than the integration of the temporal quality analyzer. This is somewhat surprising because LIVE-Qualcomm includes videos with stabilization issues as an explicit form of temporal distortions. A plausible explanation may be that  temporal distortions in LIVE-Qualcomm are strongly correlated with  spatial distortions. We treat these results as signals of the easy dataset problem, although less severe than CVD2014.

Unlike the results on CVD2014, Swin Transformer-B performs better than ResNet-50 as the spatial quality analyzer with no signal of overfitting. This is attributed to the increased number of unique scenes relative to CVD2014 (from $5$ to $54$), showing that ensuring content and distortion diversity during sample selection is necessary.  
By including all components, Model (\rom{10}) as the most ``sophisticated'' variant in our minimalistic model family matches the best performance by Li22 with long-term dependency modeling. 

\noindent \textbf{KoNViD-1k}. Models (\rom{3}) and (\rom{8}) present clear improvements over Models (\rom{1}) and (\rom{7}), respectively, and surpass VSFA, PatchVQ, and Li22 by a clear margin. This verifies the importance of quality-aware initializations for the spatial quality analyzer. Moreover, KoNViD-1k adopts a density-based fair sampling strategy to encourage content and distortion diversity, which allows the spatial quality analyzer with increased capacity to be trained without overfitting. However, the temporal information (\ie, the mean of frame-wise standard deviations
of pixel-wise frame differences) used in KoNViD-1k for sample selection appears poorly correlated with temporal content and distortion complexity, and plays little role in discouraging BIQA solutions. Moreover, the original video pool YFCC100M, from which KoNViD-1k is sampled, contains mostly legacy videos dominated by spatial distortions. These explain the marginal performance gains by adding motion features.  Last, similar to LIVE-Qualcomm, Model (\rom{10}) attains contemporary performance set by FAST-VQA.

\noindent\textbf{LIVE-VQC}. In contrast to the results on CVD2014, LIVE-Qualcomm, and KoNViD-1k, the temporal quality analyzer brings noticeable improvements (more than $10\%$ by comparing Models (\rom{4}) and (\rom{9}) to Models (\rom {1}) and (\rom{7}), respectively). This is due mainly to the manual inclusion of videos with diverse levels of motion, contributing to complex composite spatiotemporal distortions such as jittering (due to camera shake), flickering\footnote{Flickering is typically used to describe frequent and annoying luminance or chrominance changes along the temporal dimension~\cite{zeng2014characterizing}.}, and jerkiness\footnote{Jerkiness is due to temporal aliasing, in which the temporal sampling rate is low relative to the object speed, making the object appear to behave peculiarly~\cite{zeng2014characterizing}.}. Combined with quality-aware initializations, Models (\rom{6}) and (\rom{10}) experience a significant performance boost in comparison to their respective baselines.

\noindent\textbf{YouTube-UGC}. The model behaviors on YouTube-UGC resemble those on CVD2014. In particular, the two baselines, Models (\rom{1}) and (\rom{6}), perform satisfactorily. The added gains by the temporal quality analyzer are very limited (and sometimes negative),  compared to quality-aware initializations. The increased content diversity (by a density-based sampling strategy) and complexity (often due to scene changes) can be handled by a better spatial quality analyzer (\ie, Swin Transformer-B over ResNet-50). This suggests that YouTube-UGC is dominated by spatial distortions, and exhibits the easy dataset problem.  
In fact, two temporal features (\ie, the number of bits used to encode a P-frame and the standard deviation of compressed bitrates of one-second video chunks) are used for sampling selection, which turn out to be inadequate in identifying challenging videos characterized by temporal distortions.

\noindent\textbf{LBVD}. The most important observation is that the temporal quality analyzer plays an important role on LBVD, whose effect is even more pronounced than that on LIVE-VQC.  Regardless of the variations of the spatial quality analyzer (\ie, whether to adopt a more sophisticated architecture and whether to start from quality-aware initializations), the SRCC results are roughly around $0.7$. When the temporal quality analyzer is incorporated, the performance is significantly boosted to about $0.88$, uplifting by $25.7\%$. This arises because  LBVD consists of live broadcasting videos based on HTTP adaptive streaming, which are characterized by \textit{localized} temporal distortions like rebuffering, stalling, and jittering. Additionally, LBVD mainly contains low-resolution front-face videos with different backgrounds, whose spatial information is far simpler than that in other datasets. Hence, a simple spatial quality analyzer (\eg, ResNet-50) suffices to analyze spatial quality.

\noindent\textbf{LSVQ}. As the largest VQA dataset so far, LSVQ neutralizes the performance gains by the IQA pre-trained initialization. Meanwhile, LSVQ generally provides more effective initializations to improve  BVQA models on other small-scale VQA datasets. 
Nevertheless, like KoNViD-1k and YouTube-UGC, LSVQ is dominated by  spatial distortions, on which Models (\rom{1}) and (\rom{7}) achieve competitive results, weakening its role of benchmarking BVQA models. 

LSVQ includes a second test set of 1080p videos, with a resolution higher than that in the training and default test sets. We find that the performance on LSVQ-1080p degrades noticeably for two main reasons. The first is aggressive spatial downsampling, which results in  loss of fine details relevant to BVQA. As evidence, Li22 which employs the same (but fixed) spatial and temporal quality analyzers as in Models (\rom{4}) to (\rom{6}) achieves better performance on LSVQ-1080p by operating on the actual spatial resolution.
The second is that 1080p videos commonly lie in the high-quality regime, which are difficult for BVQA models to discriminate their fine-grained relative quality.

\noindent\textbf{LIVE-YT-Gaming}. The trend on LIVE-YT-Gaming is similar to that on KoNViD-1k, YouTube-UGC, and LSVQ. First, a more powerful spatial quality analyzer in terms of both initializations and network architectures is more beneficial. It is noteworthy that the videos in LIVE-YT-Gaming are rendered by computer graphics techniques. The performance gains by IQA and VQA pre-trained initializations indicate that the domain gap between gaming and natural videos is relatively small at least for the quality assessment task. This also elucidates previous practices of employing NSS \cite{yu2022subjective} in gaming VQA. Second,  the temporal quality analyzer is secondary, despite the fact that objects in gaming videos often move fast, and scene transition is also frequent (especially for first-person video games).

Putting the computational analysis results together, we make the following conclusive remarks. \underline{First}, we empirically rank the eight VQA datasets according to the severity of the easy dataset problem (from least to most): 
\begin{itemize}
    \item LBVD, that relies heavily on temporal quality analysis by including live broadcasting videos with limited spatial information;
    \item LIVE-VQC, that carefully selects videos of complex temporal distortions;
    \item LIVE-Qualcomm, that contains certain temporal distortions but strongly correlated with  spatial distortions (see also Section~\ref{subsec:cde});
    \item KoNViD-1k, YouTube-UGC, LSVQ, and LIVE-YT-Gaming, that are dominated by spatial distortions with some content diversity constraints;
    \item CVD2014, that is dominated by spatial distortions with limited content diversity.
\end{itemize} 
\underline{Second}, although the temporal quality analyzer is indispensable in LBVD and LIVE-VQC, our instantiation is embarrassingly simple. We subsample a set of video chunks (relative to the key frames), which are subject to independent local temporal analysis by a pre-trained network for action recognition. Neither end-to-end fine-tuning nor long-term dependency is implemented. This suggests that there is still ample room for constructing better VQA datasets that desire a complete temporal quality analysis.
\underline{Third}, encouraging sample diversity either manually~\cite{nuutinen2016cvd2014,ghadiyaram2017capture,sinno2018large,chen2019qoe,yu2022subjective} or automatically~\cite{hosu2017konstanz,wang2019youtube,ying2021patch}  is necessary to promote generalizable BVQA models, as evidenced by the results of switching ResNet-50 to Swin Transformer-B as the spatial quality analyzer. The missing piece during dataset construction is to simultaneously encourage sample difficulty in a way to favor more appropriate (spatio)temporal quality analyzers, and ultimately to prevent close-to-BIQA solutions.
\underline{Fourth}, albeit chasing top performance on existing VQA datasets is not our goal, Model (\rom{10}) obtains the best weighted average results, where the weighting is proportional to the number of videos in each dataset. This justifies the rationality of the decomposition and implementation of our minimalistic BVQA model family used for computational analysis.

\begin{table*}
\centering
\renewcommand{\arraystretch}{1.25}
\caption{Cross-dataset evaluation as a way of examining the second criterion of the easy dataset problem. All BVQA models are trained on LSVQ and tested on the other seven VQA datasets without fine-tuning, while BIQA models are trained on the respective IQA datasets. The best results are highlighted in bold}
\label{performance_cross_dataset_evaluation}
\resizebox{1\textwidth}{!}{
\begin{tabular}{l|c|c|c|c|c|c|c}
\toprule[.15em]
 \multirow{2}{*}{Method} & CVD2014  & LIVE-Qualcomm & KoNViD-1k & LIVE-VQC & YouTube-UGC& LBVD& LIVE-YT-Gaming\\
  & SRCC/PLCC & SRCC/PLCC  & SRCC/PLCC & SRCC/PLCC & SRCC/PLCC & SRCC/PLCC & SRCC/PLCC  \\
\hline
\rowcolor{tablegray}
\multicolumn{8}{l}{\textit{Competing BIQA Methods}} \\
 UNIQUE~\cite{zhang2021uncertainty} & 0.504/0.581 & 0.644/0.666 &  0.777/0.772 & 0.741/0.774  & 0.539/0.582 & 0.401/0.411  & 0.411/0.438   \\
 CLIP-IQA~\cite{wang2023exploring}     & 0.515/0.582 & 0.573/0.607 &  0.781/0.783 & 0.725/0.782  & 0.549/0.601 & 0.464/0.475  & 0.441/0.540   \\
 LIQE~\cite{zhang2023blind}      & 0.587/0.609 & 0.650/0.644 &  0.757/0.721 & 0.717/0.719  & 0.379/0.433 & 0.440/0.471  & 0.334/0.351   \\
 StairIQA~\cite{sun2023blind}       & 0.542/0.591 & 0.544/0.578 &  0.744/0.751 & 0.696/0.756  & 0.666/0.662 & 0.631/0.635  & 0.464/0.520   \\
 \rowcolor{tablegray}
\multicolumn{8}{l}{\textit{Competing BVQA Methods}} \\
VSFA \cite{li2019quality}  & 0.756/0.760 & 0.604/0.633 &  0.810/0.811 & 0.753/0.795  & 0.718/0.721 & 0.713/0.678  & 0.669/0.698   \\
PatchVQ \cite{ying2021patch}  & ---/--- & ---/--- & 0.791/0.795  & 0.770/0.807  & ---/--- &  ---/--- & ---/---   \\
Li22 \cite{li2022blindly}  & 0.817/0.811 &  0.732/0.744& 0.843/0.835  & 0.793/0.811  & 0.802/0.792 & \textbf{0.788}/\textbf{0.769}  & 0.690/0.740  \\
FAST-VQA \cite{wu2022fast} & 0.805/0.814 & \textbf{0.737}/\textbf{0.749} & 0.859/0.856  & \textbf{0.822}/\textbf{0.845}  &  0.730/0.747& 0.734/0.715 & 0.620/0.666   \\
\hline
\rowcolor{tablegray}
\multicolumn{8}{l}{\textit{Proposed Methods}} \\
Model (\rom{4})  & 0.780/0.802 & 0.533/0.590 & 0.826/0.820  & 0.749/0.789  & 0.802/0.806 &  0.674/0.661 & 0.656/0.717   \\
Model (\rom{9})  & \textbf{0.858}/\textbf{0.867} & 0.635/0.699 & \textbf{0.889}/\textbf{0.887}  &  0.798/0.837 & \textbf{0.826}/\textbf{0.833} & 0.670/0.655  & \textbf{0.711}/\textbf{0.755}  \\
\bottomrule[.15em]
\end{tabular}
}
\end{table*}

\subsection{Auxiliary Computational Analysis}\label{subsec:cde}
To support our analysis results, we continue to examine the second criterion that characterizes the easy dataset problem by testing the generalization of BVQA models trained on LSVQ to the other seven datasets. The competing models include four BIQA methods, UNIQUE~\cite{zhang2021uncertainty}, CLIP-IQA~\cite{wang2023exploring}, LIQE~\cite{zhang2023blind} and StairIQA~\cite{sun2023blind}, and six BVQA methods, VSFA, PatchVQ, Li22, FAST-VQA, Model (\rom{4}) and Model (\rom{9}). UNIQUE and LIQE are both trained on six IQA datasets, including LIVE~\cite{sheikh2006statistical}, CSIQ~\cite{larson2010most}, KADID-10k~\cite{lin2019kadid}, BID~\cite{ciancio2011no}, CLIVE~\cite{ghadiyaram2015massive}, and KonIQ-10k~\cite{hosu2020koniq}. CLIP-IQA is fine-tuned on KonIQ-10k~\cite{hosu2020koniq}. StairIQA is trained on five in-the-wild IQA datasets consisting of BID~\cite{ciancio2011no}, CLIVE~\cite{ghadiyaram2015massive}, KonIQ-10k~\cite{hosu2020koniq}, SPAQ~\cite{fang2020perceptual}, and FLIVE~\cite{ying2020patches}.
The experimental results are listed in Table \ref{performance_cross_dataset_evaluation}, from which we make similar observations. First, trained on the large-scale LSVQ dominated by spatial distortions, Models (\rom{4}), and (\rom{9}), even equipped with the temporal quality analyzer, generalize poorly to LBVD and LIVE-Qualcomm, characterized by temporal distortions. Combining with the intra-dataset results, we reinforce the claim that the spatial and temporal distortions in LIVE-Qualcomm are highly correlated.
In the intra-dataset setting, such correlation can be effectively exploited; however in the cross-dataset setting, the models may have to capture  temporal distortions to enable generalization on LIVE-Qualcomm, and such capability is less likely induced by LSVQ.
Second, for CVD2014, KoNViD-1k, and YouTube-UGC, both Model (\rom{4}) and Model (\rom{9}) generalize reasonably, and a more powerful spatial quality analyzer gives better performance, pointing out the severity of the easy dataset problem. 
Third, for LIVE-VQC whose problem is more severe than LBVD but less severe than KoNViD-1k, Model (\rom{4}) performs marginally, and Model (\rom{9}) performs moderately, aligning well with the intra-dataset results.
Fourth, although LIVE-YT-Gaming is analyzed to be spatial-distortion-dominant, the model generalizability is weaker compared to that of CVD2014, KoNViD-1k, and YouTube-UGC. We attribute this to the domain gap between gaming and natural videos, which can not be ignored in the cross-dataset setting.

For BIQA models, we observe that despite not being trained on any of the VQA datasets, they deliver reasonable performance in evaluating video quality. For example, UNIQUE well handles LIVE-Qualcomm, KoNViD-1k, and LIVE-VQC, while StairIQA performs good on YouTube-UGC and LBVD. We ascribe the results to three main factors. First, these BIQA models are trained on multiple IQA datasets or leverage powerful feature extraction modules (\eg,~CLIP). Second, the domain gap between certain VQA datasets (\eg,~LIVE-Qualcomm, KoNViD-1k, and LIVE-VQC) and training IQA datasets is relatively small, which is further supported by the practice of using pre-trained weights on IQA datasets as initializations for BVQA. Third, spatial quality analysis is dominant on these VQA datasets, which aligns with the easy dataset problem exposed through our main computational analysis.

\section{Further Ablation Studies}
In the previous section, we have identified the easy dataset problem that occurs in existing VQA datasets. In this section, we accompany our computational analysis with comprehensive ablation experiments, investigating additional BVQA design choices pertaining to the basic building blocks. These include 1) the spatial resolution and temporal downsampling rate in the video preprocessor, 2) the backbone, initialization, and feature representation in the spatial quality analyzer, and 3) the feature aggregation and score pooling in the quality regressor. We also ablate the loss function as the key part of the optimization.  We report the results of Model (\rom{1}) on LSVQ unless otherwise specified.

\begin{figure*}[htbp]
\centering
\subfigure[LSVQ]
{\label{fig:resolution_lsvq_test}
\includegraphics[width=0.225\textwidth]{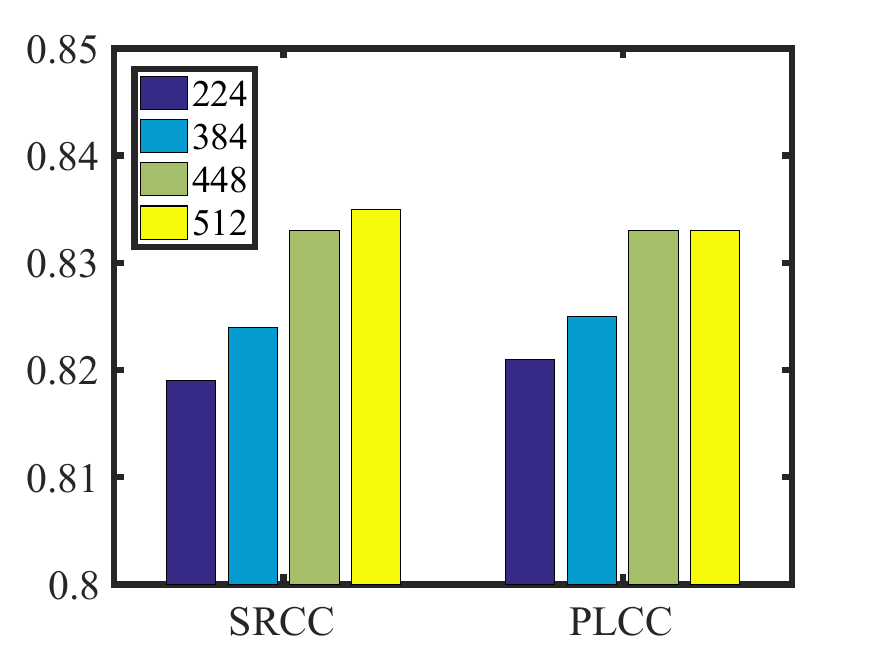}
}
\subfigure[LSVQ-1080p ]
{\label{fig:resolution_lsvq_test_1080p}
\includegraphics[width=0.225\textwidth]{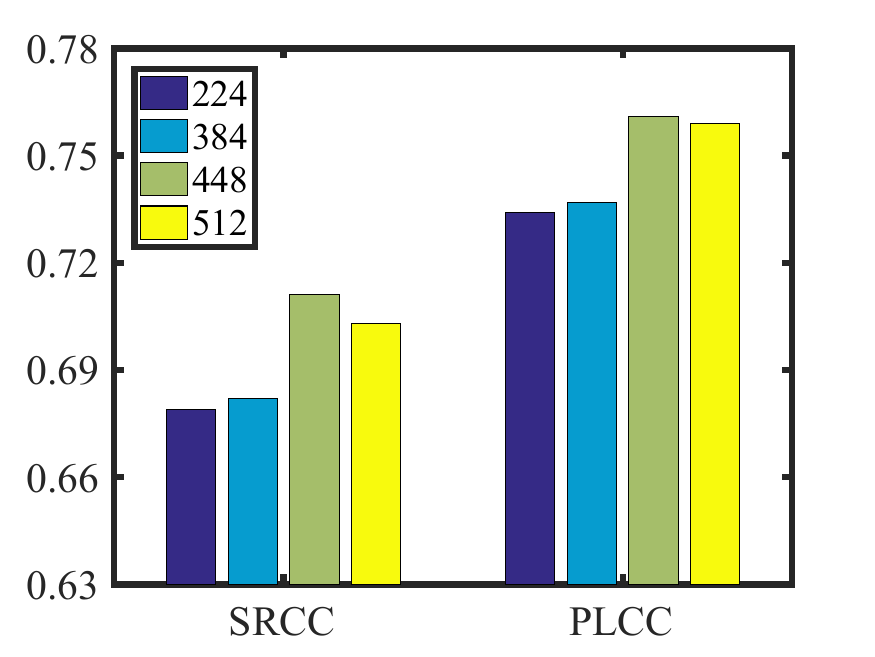}
}
\subfigure[LSVQ ]
{\label{fig:sampling_lsvq_test}
\includegraphics[width=0.225\textwidth]{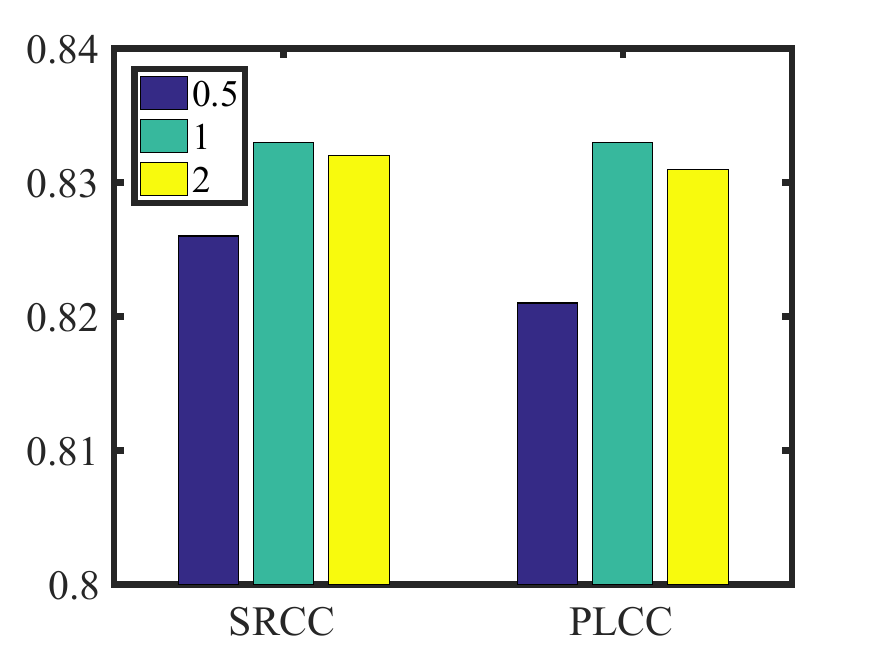}
}
\subfigure[LSVQ-1080p ]
{\label{fig:sampling_lsvq_test_1080p}
\includegraphics[width=0.225\textwidth]{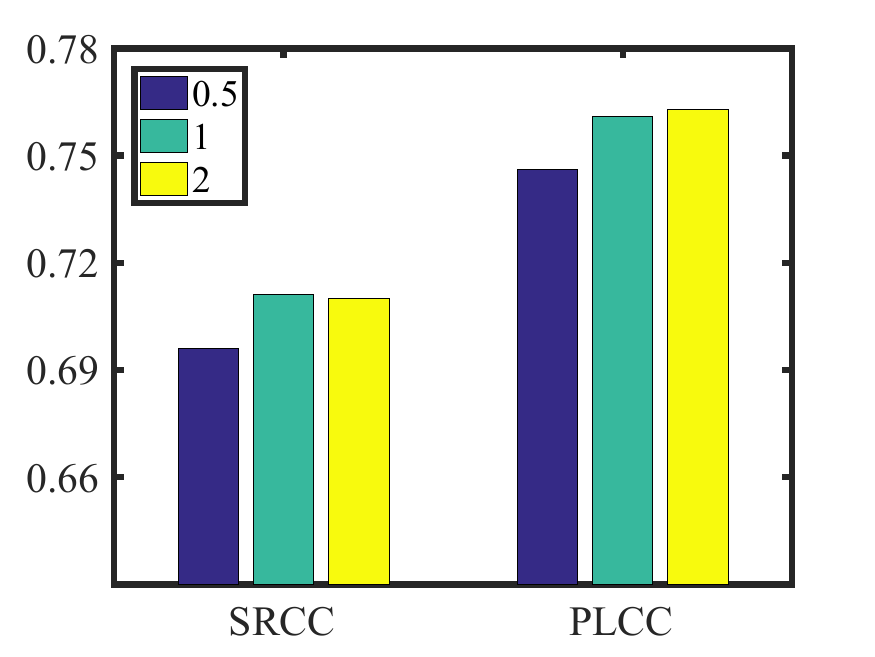}
}
\caption{Performance on the LSVQ test sets by varying different spatial resolutions and temporal sampling rates for key frames under Model (\rom{1}). The legends in (a) and (b) represent the input spatial resolutions, and the legends in (c) and (d) represent the temporal sampling rate for key frames.
}
\label{fig:resolution_sampling}
\end{figure*}

\begin{figure}[t]
\centering
\subfigure[LSVQ]
{
\includegraphics[width=0.225\textwidth]{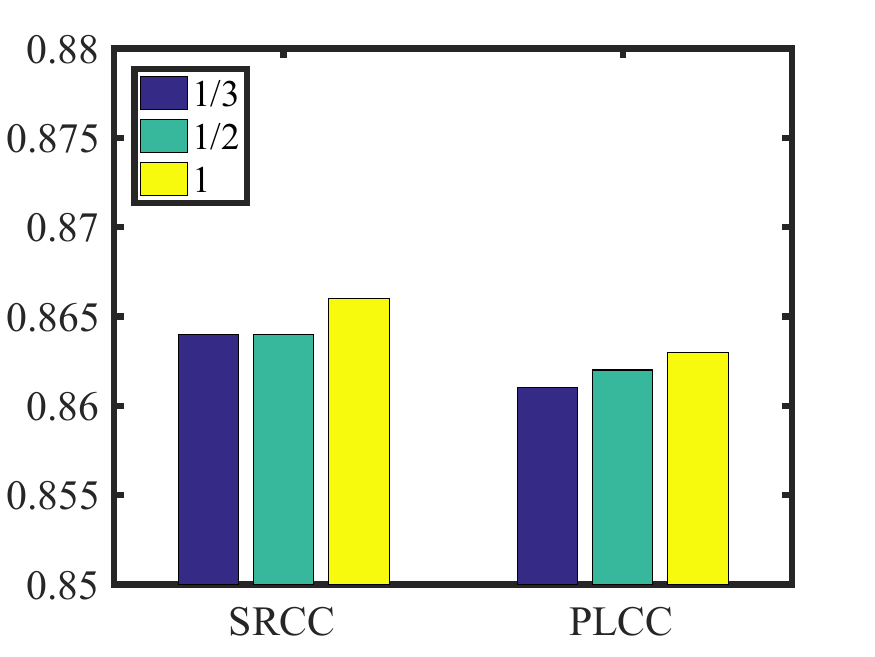}
}
\subfigure[LSVQ-1080p ]
{
\includegraphics[width=0.225\textwidth]{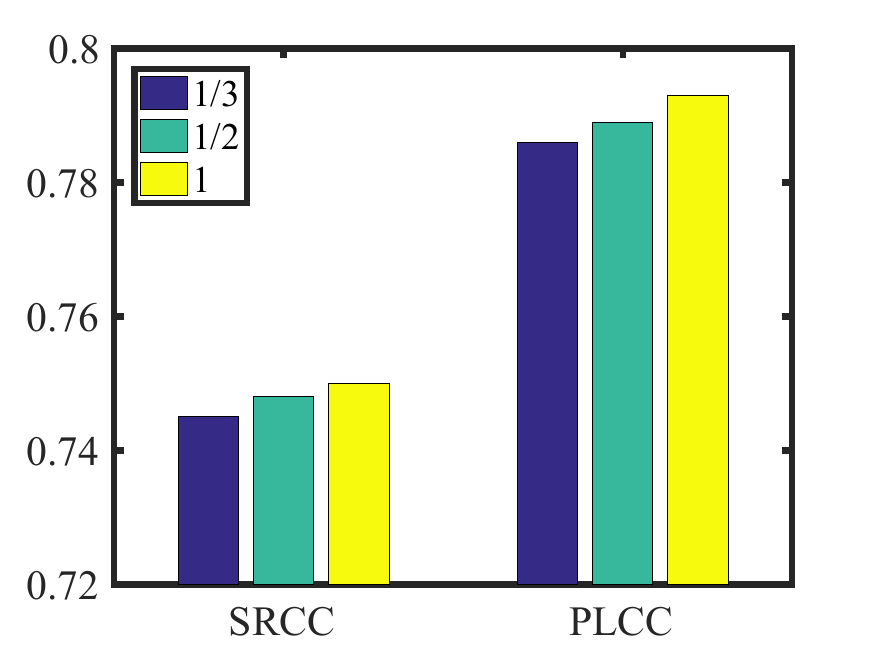}
}
\caption{Performance on the LSVQ test sets by varying different temporal sampling rates (in the legends) for video chunks under Model (\rom{4}). 
}
\label{fig:samplint_rate_video_chunk}
\end{figure}

\subsection{Video Preprocessor}

\noindent\textbf{Spatial Resolution for Key Frames}. 
The spatial resolution, $L_s$, in our video preprocessor determines how aggressively we choose to downsample each key frame. In our default settings, $L_s = 448$ for the ResNet-50, and $L_s = 384$ for the Swin Transformer-B. Here we make other choices: $\{224,384, 448,512\}$, whose results are shown in Fig. \ref{fig:resolution_lsvq_test} and \ref{fig:resolution_lsvq_test_1080p}.  We find that the performance is fairly stable for different spatial resolutions,  suggesting that spatial distortions presented in current VQA datasets are not sensitive to spatial downsampling. Although operating at the actual resolution may be advantageous (by comparing Li22~\cite{li2022blindly} to Models (\rom{4}) to (\rom{6})), it is computationally prohibitive in the case of end-to-end training.

\noindent\textbf{Temporal Sampling Rate for Key Frames}.
Natural videos are highly redundant in the temporal dimension. We reduce the temporal sampling rate to $R_a=1$ fps, from which we extract quality-aware spatial features. Here, we explore a set of alternative choices: $\{0.5,1,2\}$, and illustrate the results in Fig. \ref{fig:sampling_lsvq_test} and \ref{fig:sampling_lsvq_test_1080p}. When the temporal sampling rate is extremely low (\eg, $0.5$ fps), the performance drops considerably. Increasing the temporal sampling rate from $1$ to $2$ fps has little impact on quality prediction, but doubles the computation. Overall, $R_a = 1$ is a decent choice for  BVQA models to work well on existing VQA datasets.

\noindent\textbf{Temporal Sampling Rate for Video Chunks}.
We explore the impact of the temporal sampling rate, $R_b$, for video chunks. Specifically, we sample $R_b$ from $\{1/3, 1/2, 1\}$ under Model (\rom{4}), whose results are listed in Fig. \ref{fig:samplint_rate_video_chunk}. We observe that setting $R_b$ to slightly lower values induces little performance variations. Altogether, $R_a$ and $R_b$ give a rough but convenient characterization of the temporal content and distortion complexity. A good VQA dataset shall not permit the setting of too low $R_a$ and $R_b$ (\eg, those used in our minimalistic BVQA model family), which corresponds to very sparsely localized temporal analysis.

\begin{table}
\centering
\renewcommand{\arraystretch}{1.25}
\caption{Performance on the LSVQ test sets by varying  different backbones of the spatial quality analyzer under Model (\rom{1})}
\label{performance_backbone_lsvq}
\begin{tabular}{cc|cc|cc}
\toprule[.15em]
 \multirow{2}{*}{Backbone}& \multirow{2}{*}{Scale} & \multicolumn{2}{c|}{LSVQ} & \multicolumn{2}{c}{LSVQ-1080p}  \\
  & & SRCC  & PLCC & SRCC & PLCC  \\
\hline
\multirow{2}{*}{ResNet}& 50&  0.819         & 0.821  & 0.679  &  0.734      \\
& 200&   0.816   &  0.817 &  0.662 &   0.727 \\
\hline
\multirow{2}{*}{ViT}&Tiny &    0.815      & 0.819 &  0.685 &   0.744      \\
& Base&     0.848     & 0.849 & 0.718  &   0.769     \\
\hline
\multirow{2}{*}{ConvNeXt}&Tiny &    0.833      & 0.835 & 0.712  &    0.764    \\
&Base &     0.843     & 0.844 & 0.720  &   0.769     \\
\hline
\multirow{2}{*}{\makecell[c]{Swin \\Transformer}}& Tiny&   {0.842}       & {0.844} & {0.718}  &   {0.766}     \\
&Base &      \textbf{0.852}    & \textbf{0.854} & \textbf{0.736}  &   \textbf{0.780}     \\
\bottomrule[.15em]
\end{tabular}
\end{table}

\subsection{Spatial Quality Analyzer}

\noindent\textbf{Backbone}.
We test four backbone networks, each with two scales, to probe the impact of different network architectures and model scales on  BVQA. 
The first are ResNet-50~\cite{he2016deep}, ViT-Tiny~\cite{dosovitskiy2020image}, ConvNeXt-tiny~\cite{liu2022convnet}, and Swin Transformer-Tiny~\cite{liu2021swin} with FLOPS\footnote{The FLOPS of backbones are calculated using a single frame with the spatial resolution of $224\times224$.} around $4.5\times10^9$. The second are ResNet-200, ViT-Base, ConvNeXt-Base, and Swin Transformer-Base with FLOPS around $15.0 \times 10^9$. The spatial resolution for key frames is set to $L_s = 224$, and other configurations are the same as in Model (\rom{1}). The results are listed in Table \ref{performance_backbone_lsvq}, from which we make two useful observations. First, except for ResNets, the performance improves as the FLOPS of the backbones increase, by an average of $2.2\%$. ResNet-200 gives slightly worse results than ResNet-50, showing that scaling up ResNets for the BVQA task is more challenging. 
Second, the performance of more advanced backbones, such as Swin Transformers, is better than DNNs with standard convolution (\eg, ConvNeXt).

\begin{table}
\centering
\renewcommand{\arraystretch}{1.25}
\caption{Performance on the LSVQ test sets by varying different initializations under models without the temporal quality analyzer }
\label{performance_initilization_lsvq}
\resizebox{0.48\textwidth}{!}{
\begin{tabular}{cc|cc|cc}
\toprule[.15em]
 \multirow{2}{*}{Backbone}&\multirow{2}{*}{Initialization} & \multicolumn{2}{c|}{LSVQ} & \multicolumn{2}{c}{LSVQ-1080p}  \\
  & & SRCC  & PLCC & SRCC & PLCC  \\
\hline
\multirow{3}{*}{ResNet-50}&ImageNet-1k&0.819&0.821&0.679&0.734\\
&MoCoV2&0.825&0.828&0.697&0.758\\
&CLIP&0.842&0.846&0.731&0.769\\
\hline
\multirow{3}{*}{ViT-B}&ImageNet-1k&0.848&0.849&0.718&0.770\\
&MAE&0.846&0.849&0.730&0.770\\
&CLIP&\textbf{0.863}&
\textbf{0.865}&
\textbf{0.751}&
\textbf{0.802}\\
\hline
\multirow{2}{*}{\makecell[c]{Swin \\Transformer-B}}&ImageNet-1k&0.852&0.854&0.736&0.780\\
&ImageNet-22k&0.857&0.860&0.741&0.793\\
\bottomrule[.15em]
\end{tabular}
}
\end{table}

\noindent\textbf{Initialization}.\label{subsubsec:oi}
We have investigated the effectiveness of quality-aware initializations for  BVQA. Here, we further explore other possibilities: 1) initializations from pre-training on larger computer vision datasets (\eg, ImageNet-22k) and 2) initializations from unsupervised pre-training, including MoCo~\cite{he2020momentum}, MAE \cite{he2022masked}, and CLIP \cite{radford2021learning}. Specifically, we initialize ResNet-50 with weights supplied by MoCoV2 and CLIP, ViT-B by MAE and CLIP, and Swin Transformer-B by ImageNet-22k.
The spatial resolution is set to $224$, while keeping other configurations the same as in Model (\rom{1}). 

The results are given in Table \ref{performance_initilization_lsvq}, from which we have several interesting observations. First, switching from  ImageNet-1k to ImageNet-22k leads to about $1\%$ improvement on LSVQ in terms of SRCC. Second,
 we find surprisingly that MoCoV2 as a kind of contrastive learning method improves BVQA compared with its counterpart pre-trained on ImageNet-1K. MoCoV2 tries to identify different positive pairs derived from one image by a bag of data augmentation methods, some of which are quality non-preserving (\eg, color jittering and color-to-gray conversion).  We conjecture that these augmentations as visual distortions are not common in LSVQ, and thus may not hurt the quality prediction performance as measured on LSVQ.  MAE attempts to reconstruct the masked patches of an image as a way of encouraging high-level distortion-unaware features, hence not beneficial to the BVQA task.
CLIP, as a visual-language pre-training method, consistently improves  BVQA, which  aligns well with the recent studies\cite{wang2023exploring} that exploit CLIP features for BVQA  without  fine-tuning.

\begin{table}
\centering
\renewcommand{\arraystretch}{1.25}
\caption{Performance on the LSVQ test sets by varying different feature extraction strategies under Model (\rom{1}). AVG and STD stand for global average and standard deviation pooling, respectively. ``+'' means feature concatenation. Single-stage and multi-stage indicate that the feature maps are extracted from the last stage and the last three stages of ResNet-50, respectively}
\label{performance_statistics_features_resnet50_lsvq}
\begin{tabular}{cc|cc|cc}
\toprule[.15em]
\multicolumn{2}{c|}{\multirow{2}{*}{Feature Extraction}} & \multicolumn{2}{c|}{LSVQ} & \multicolumn{2}{c}{LSVQ-1080p}  \\
 & & SRCC  & PLCC & SRCC & PLCC  \\
\hline
\multirow{2}{*}{Single-stage}& AVG &   0.833  & 0.833  & 0.711  & 0.761   \\
&AVG+STD &   0.831   & 0.834  & 0.693  & 0.752  \\
\hline
\multirow{2}{*}{Multi-stage}&AVG&0.838&0.840&0.712&0.761\\
&AVG+STD&\textbf{0.839}&\textbf{0.841}&\textbf{0.720}&\textbf{0.766}\\
\bottomrule[.15em]
\end{tabular}
\end{table}

\noindent\textbf{Feature Extraction Strategy}.
The global average pooling of the last-stage feature maps is the most common feature extraction strategy in BVQA. Nevertheless, other options exist, such as 1) the global average and standard deviation pooling \cite{li2019quality} to deal with non-uniformity of spatial distortions and 2) multi-stage feature fusion to enrich the feature expressiveness. Here, we further explore the combinations of these two more involved feature extraction strategies, and show the results in Table \ref{performance_statistics_features_resnet50_lsvq}. Unlike previous studies~\cite{li2019quality,li2022blindly}, we do not observe significant improvements compared to the global average pooling of the last-stage feature maps. We believe this arises because of the end-to-end training, which encourages better quality-aware features to be learned without additional feature engineering.

\begin{table}
\centering
\renewcommand{\arraystretch}{1.25}
\caption{Performance on the LSVQ test sets by varying different feature aggregation methods under Model (\rom{1})}
\label{performance_regressor_resnet50_lsvq}
\begin{tabular}{c|cc|cc}
\toprule[.15em]
\multirow{2}{*}{Regressor}  & \multicolumn{2}{c|}{LSVQ} & \multicolumn{2}{c}{LSVQ-1080p}  \\
   & SRCC  & PLCC & SRCC & PLCC  \\
\hline
MLP &      0.833  & 0.833 &  0.711 & 0.761    \\
GRU &  0.831    & 0.829 &  0.687 & 0.746  \\
Transformer &   \textbf{0.841}   & \textbf{0.841} & \textbf{0.717}  & \textbf{0.765}  \\
\bottomrule[.15em]
\end{tabular}
\end{table}

\subsection{Quality Regressor}
\noindent\textbf{Temporal Feature Aggregation}.
In our minimalistic BVQA model family, we adopt an FC layer to regress local quality scores. We may as well employ sequential models such as GRU \cite{cho2014properties}, and the Transformer \cite{vaswani2017attention} to capture long-term dependencies among key frames or video chunks under Model (\rom{1}). The results are listed in Table \ref{performance_regressor_resnet50_lsvq}, from which we observe that FC outperforms GRU. The Transformer achieves the best performance, but the improvement relative to FC is marginal, indicating that long-term dependencies are relatively weak in LSVQ.

\begin{table}
\centering
\renewcommand{\arraystretch}{1.25}
\caption{Performance on the LSVQ test sets by varying different temporal pooling strategies under Model (\rom{1})}
\label{performance_quality_pooling_resnet50_lsvq}
\begin{tabular}{c|cc|cc}
\toprule[.15em]
\multirow{2}{*}{Pooling Strategy}  & \multicolumn{2}{c|}{LSVQ} & \multicolumn{2}{c}{LSVQ-1080p}  \\
  & SRCC  & PLCC & SRCC & PLCC  \\
\hline
Simple average &         0.833  & 0.833  &  0.711 & \textbf{0.761}       \\
Temporal hysteresis &   0.826   & 0.825  &  0.689 & 0.738   \\
Learned  &   \textbf{0.837}   & \textbf{0.835}  &  \textbf{0.715} & 0.757   \\
\bottomrule[.15em]
\end{tabular}
\end{table}

\noindent\textbf{Temporal Pooling}.
The temporal pooling merges frame-level or chunk-level quality scores into a video-level quality score. Representative strategies include simple average pooling, worst-quality-based pooling \cite{park2012video}, temporal variation pooling \cite{ninassi2009considering}, and temporal hysteresis pooling \cite{seshadrinathan2011temporal,li2019quality}. Considering that only differentiable pooling methods can be integrated for end-to-end optimization, we employ three pooling strategies under Model (\rom{1}): the simple average pooling, the temporal hysteresis pooling, and the learned pooling using a 1D convolution layer (with a kernel size of $5$). The results are listed in Table \ref{performance_quality_pooling_resnet50_lsvq}, where we find that the temporal hysteresis pooling and the learned pooling are no better than the simple average pooling. This shows that the videos included in LSVQ are not capable of giving credit to BVQA methods with explicit temporal modeling.

\begin{table}
\centering
\renewcommand{\arraystretch}{1.25}
\caption{Performance on the LSVQ test sets by varying different loss functions under Model (\rom{1})}
\label{performance_loss_function_resnet50_lsvq}
\begin{tabular}{c|cc|cc}
\toprule[.15em]
\multirow{2}{*}{Loss} & \multicolumn{2}{c|}{LSVQ} & \multicolumn{2}{c}{LSVQ-1080p}  \\
  & SRCC  & PLCC & SRCC & PLCC  \\
\hline
L1 &   0.825   & 0.825  &  0.693 & 0.746   \\
L2 &    0.822  & 0.823 &  0.686 & 0.741 \\
SRCC &   \textbf{0.833}  & 0.825  &  0.697 & 0.749  \\
PLCC &   \textbf{0.833}  & \textbf{0.833}  &  \textbf{0.711} & \textbf{0.761} \\
\bottomrule[.15em]
\end{tabular}
\end{table}

\subsection{Loss Function}
BVQA is commonly formulated as a regression problem, optimized for MAE or MSE as the loss function. Here we compare MAE and MSE with the default PLCC. We also include a differentiable version of SRCC~\cite{li2022blindly} in Table \ref{performance_loss_function_resnet50_lsvq}. It is clear that PLCC performs the best, especially on LSVQ-1080p, which justifies its use in optimizing our minimalistic BVQA model family. Nevertheless, BVQA models optimized for PLCC suffer from scale ambiguity, meaning that the results stay unchanged if we scale the model prediction $\hat{q}$ by a positive scalar $\alpha$ to $\alpha\hat{q}$. Thus, same as using PLCC for evaluation~\cite{antkowiak2000final}, to obtain more interpretable quality predictions, a monotonic logistic function that maps model predictions to MOSs can be fitted.

In summary, by ablating more BVQA design choices on LSVQ, we arrive at similar observations in support of the main computational analysis: the BVQA performance benefits from more advanced spatial quality analysis, but not from temporal quality analysis.

\section{Discussion}
To reliably measure the progress of BVQA and to develop better BVQA methods, it is crucial to construct next-generation VQA datasets that overcome the easy dataset problem. Here we propose to revisit the two key steps in subjective VQA: sample selection and subjective testing.

From the sample selection perspective, the selected videos should be \textit{diverse} in terms of content and distortion types and \textit{difficult} in terms of falsifying existing BVQA models. To promote diversity, Vonikakis \etal~\cite{vonikakis2016shaping} described an elegant computational procedure that is mathematically grounded. The tricky part lies in the specification of video attributes, which should be efficiently computed and perceptually relevant (especially those related to time-varying video quality). To promote difficulty, Wang \etal~\cite{wang2021troubleshooting,wang2021active} proposed to leverage the group maximum differentiation (gMAD) competition~\cite{ma2018group} to let a group of BIQA methods compete with one another on a large-scale unlabeled image dataset. In a similar spirit, we may select a group of BVQA models (including those from our minimalistic BVQA model family) to participate in the gMAD competition on a large-scale unlabeled video dataset. The resulting video subset has great potential in falsifying the competing BVQA models. Alternatively, we may borrow ideas from the active learning literature, and co-train an auxiliary failure prediction module~\cite{cao2024image} (along with the main BVQA model) to predict sample difficulty. In general, sample selection is an NP-hard problem~\cite{davis1997adaptive} unless the diversity and difficulty measures have special structures to be
exploited.

From the subjective testing perspective, instead of relying on absolute category rating that gives a single scalar to indicate perceptual video quality, we may resort to single stimulus continuous quality rating~\cite{series2012methodology}, in which a test video is rated continuously over time, giving rise to a quality curve over time. Such quality annotations force BVQA models to analyze temporal quality as a direct means of preventing close-to-BIQA solutions. However,  continuous quality rating has notorious drawbacks regarding rating efficiency and reliability. Following this line of thought,  we may formulate different variants of BVQA problems that emphasize temporal quality assessment: 1) localization of worst and best quality regions in space and time, and  2) two-alternative forced choice between two (non-overlapping) video chunks with better perceived quality.

\section{Conclusion}
We have conducted a computational analysis of eight VQA datasets by designing minimalistic BVQA models, which perform aggressive spatiotemporal downsampling as preprocessing and rely on as little temporal quality analysis and aggregation as possible. Despite being minimalistic,  our models achieve very competitive quality prediction performance compared to existing BVQA methods. We view the results not as the advancement of BVQA, but as the underlying problem of existing VQA datasets---the easy dataset problem. We empirically rank the eight VQA datasets according to the severity of the easy dataset problem, which is supported by the cross-dataset evaluation and ablation experiments. We hope our computational studies shed light on how to construct next-generation VQA datasets and models in a closed loop.


%





\ifCLASSOPTIONcaptionsoff
  \newpage
\fi



%



\bibliographystyle{IEEEtran}
\bibliography{bare_adv}

%








\end{document}